\documentclass[10pt,twocolumn,letterpaper]{article}

\usepackage{cvpr}              

\usepackage{algorithm}
\usepackage{algorithmic}
\usepackage{booktabs}

\newcommand{\smed}{S_{\text{med}}}
\newcommand{\mad}{\sigma_{\text{med}}}
\newcommand{\st}{S_{95}}
\newcommand{\sclam}{S_{\text{ROSS}}}

\usepackage{xcolor}   
\usepackage{colortbl} 
\definecolor{Gray}{gray}{0.9}
\definecolor{LightCyan}{rgb}{0.88,1,1} 

\usepackage{subcaption}
\usepackage{graphicx}

\definecolor{cvprblue}{rgb}{0.21,0.49,0.74}
\usepackage[pagebackref,breaklinks,colorlinks,allcolors=cvprblue]{hyperref}
\usepackage[accsupp]{axessibility}  

\def\paperID{} 
\def\confName{CVPR}
\def\confYear{2026}

\title{A Robust Out-of-Distribution Detection Framework via Synergistic Smoothing}

\author{Maria Stoica\\
Imperial College London\\
{\tt\small m.stoica22@imperial.ac.uk}
\and
Abdelrahman Hekal\\
Zeroth Research\\
{\tt\small abdelrahman.hekal@zeroth.org}
\and
Alessio Lomuscio\\
Imperial College London\\
{\tt\small a.lomuscio@imperial.ac.uk}
}

\begin{document}
\maketitle
\begin{abstract}
Reliable out-of-distribution (OOD) detection is a critical requirement for the safe deployment of machine learning systems. 
Despite recent progress, state-of-the-art OOD detectors are highly susceptible to adversarial attacks, which undermines their trustworthiness in automated systems. 
To address this vulnerability, we apply median smoothing to baseline OOD detection scores, balancing clean and adversarial accuracies. 
Our key insight is that the noisy samples generated for median smoothing can be repurposed to quantify the local instability of the base score. 
We observe that OOD samples exhibit higher instability under perturbation. 
Based on this, we propose ROSS, a novel and robust post-hoc OOD detector that leverages the instability of baseline scores to further distinguish between in-distribution (ID) and OOD samples. 
ROSS achieves symmetric robustness, performing strongly against both score-minimising and score-maximising attacks, unlike prior work.
This symmetric defence leads to state-of-the-art robustness, outperforming prior methods by up to 40 AUROC points.
We demonstrate ROSS's effectiveness on extensive experiments across CIFAR-10, CIFAR-100, and ImageNet.
Code is available at: \url{https://github.com/Abdu-Hekal/ROSS}.
\end{abstract}    
\section{Introduction}\label{sec:intro}
\begin{figure}[h!]
    \centering
        \includegraphics[width=0.95\linewidth]{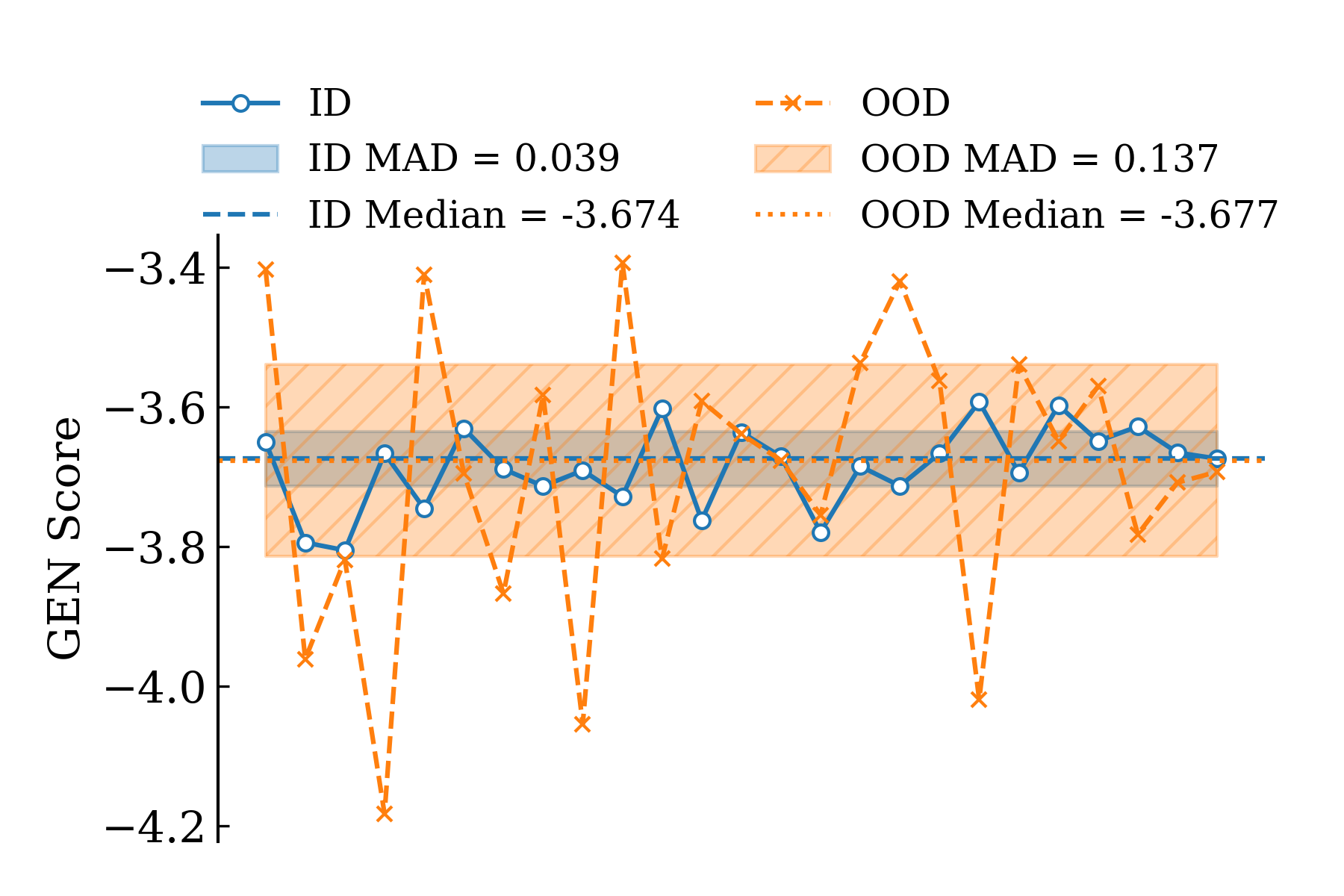}
    \caption{
    Comparison of GEN score profiles for an ID sample (CIFAR-10) and an OOD sample (Tiny ImageNet) under input perturbations (25 noisy variants). Despite similar medians, the OOD sample shows much higher instability (MAD), illustrating how instability metrics help distinguish ID from OOD when base scores overlap.
    }        
    \label{fig:enter-label}
\end{figure}
As deep neural networks (DNNs) are increasingly integrated into a wide range of applications, from autonomous vehicles and medical diagnostics to financial and social systems, their robustness and reliability are of utmost importance.
Despite achieving high accuracy on in-distribution (ID) data, DNNs remain highly susceptible to adversarial perturbations and other forms of distribution shift, leading to confident yet incorrect predictions.

While pre-deployment strategies, such as design-time verification~\cite{liu_algorithms_2021}, provide an essential first layer of safety, they cannot anticipate the full spectrum of adversarial conditions or unseen data that arise in dynamic real-world environments.   
Thus, ensuring that models can maintain reliable performance, or at least recognise when they are being manipulated or operating out of distribution, becomes a core challenge for achieving robust, trustworthy AI systems.

To address the challenge of differentiating between in-distribution
(ID) and out-of-distribution (OOD) data, researchers have proposed numerous methods for detection.
Within this landscape, our
investigation is centred on \textit{post-hoc} detectors.  
The key advantage of these methods is their applicability to pre-trained
models, making them a flexible and practical choice.

OOD detection
methods~\cite{amodei_concrete_2016,hendrycks_baseline_2017} have
emerged as a critical safeguard, providing real-time awareness
needed to identify and handle novel inputs before they can lead to unsafe outcomes.  
However, the existence of an OOD detector alone does not guarantee safety; its robustness under adversarial conditions directly determines the trustworthiness of the overall system.  
Recent studies~\cite{azizmalayeri2022your,lorenz_deciphering_2024} demonstrate that many existing OOD detectors are themselves vulnerable to adversarial attacks, where a malicious actor can craft an input that is OOD but is classified as in-distribution. 
Such attacks effectively disable the system's first line of defence, posing a significant risk to safety-critical applications.

Consequently, developing OOD detectors that are resilient to adversarial manipulation is a key step toward building genuinely robust AI systems.
To this end, we propose ROSS: a \textbf{R}obust \textbf{O}OD Detector via \textbf{S}ynergistic \textbf{S}moothing, a post-hoc OOD detector that combines median smoothing with score-based confidence estimation to achieve enhanced resilience against adversarial perturbations, outperforming state-of-the-art OOD detectors. 

Our method can be applied to any state-of-the-art OOD scoring technique such as MSP~\cite{hendrycks_baseline_2017},
Energy~\cite{liu_energy-based_2020}, and GEN~\cite{liu_gen_2023}.
For a given input, ROSS generates a set of $n$ noisy samples and repurposes them to compute two key statistics grounded in robust statistics:
\begin{enumerate}
    \item A robust central tendency: the \textbf{Median Score} ($\smed$).
    \item A robust measure of instability: the \textbf{Median Absolute Deviation} ($\mad$).
\end{enumerate}
Because this computation is non-directional and gradient-free, it is not predisposed to any single attack vector. 
As a result, ROSS achieves symmetric robustness, performing strongly and consistently against both score-minimising (PGD-min) and score-maximising (PGD-max) attacks.

The stability of a model's OOD score under input perturbation can serve as a powerful signal for OOD detection~\cite{dcbe7abf4db64d1b89bf9802585660ed, chen_leveraging_2025}. The principle is that score stability directly reflects the model's epistemic uncertainty. 
For an ID sample, the model's learned function is smooth and well-defined, yielding small perturbations that produce a stable, low-variance set of scores, a sign of high confidence. 
Conversely, an OOD sample inhabits an uncharted region of the input space where the model behaves erratically. There, the same perturbations result in a high-variance spread of scores, signifying low confidence and high uncertainty.

While score stability is a valuable cue, it is not sufficient on its own. A reliable OOD detector must also consider the absolute magnitude of the score — that is, how confidently the model classifies a sample as ID or OOD. Stability without a strong baseline score can lead to ambiguous or misleading detections, especially near decision boundaries.

In this work, we show that a fundamental property that distinguishes OOD data is the inherent instability of its score landscape. 
Existing methods, such as PRO~\cite{chen_leveraging_2025} and ODIN~\cite{dcbe7abf4db64d1b89bf9802585660ed}, indirectly leverage this property, albeit with opposing heuristics which lead to asymmetric robustness against attacks. 
PRO applies perturbations to reduce the model's confidence, whereas ODIN does the opposite, perturbing inputs to increase confidence.

Our key insight is that this directional optimisation is an insufficient proxy for robustness. 
We instead propose directly quantifying the instability of the local score landscape and using it as a confidence multiplier, amplifying or suppressing the base OOD score in a way that reflects both the score's magnitude and its local robustness, thereby creating an OOD score that is robust against maximising and minimising attacks.

In summary, our contributions are as follows:
\begin{itemize}
    \item We are the first to identify and analyse the asymmetric vulnerability of existing perturbation-based OOD detectors, such as PRO and ODIN, which defend against one attack direction but remain vulnerable to the other.
    \item We propose \textbf{ROSS}, a robust, model-agnostic post-hoc OOD detection framework that enhances existing OOD scores through localised stability and confidence-aware thresholding, and evaluate it extensively across CIFAR-10, CIFAR-100~\cite{krizhevsky_learning_2009}, and ImageNet~\cite{deng_imagenet_2009}.
    \item We conduct a detailed analysis to isolate the contributions of stability and confidence in OOD detection, demonstrating that both components are critical for effectively distinguishing ID and OOD data.
    \item We evaluate ROSS under adversarial perturbations on CIFAR-10 and CIFAR-100, demonstrating that it consistently improves robustness relative to baseline detectors.
\end{itemize}

The rest of the paper is organised as follows.
In Section~\ref{sec:rel_work}, we provide a brief discussion of related works.
We then introduce the necessary background on DNNs, OOD detection, and adversarial robustness in Section~\ref{sec:preliminaries}.
In Section~\ref{sec:methodology}, we present ROSS and explain how we utilise median smoothing to generate a robust OOD score.
Then, in Section~\ref{sec:evaluation}, we evaluate ROSS against commonly used baselines and datasets, including results on robust DNNs and an assessment of ROSS against adversarial attacks.
\section{Related Work}\label{sec:rel_work}
Our research builds upon two primary domains: post-hoc out-of-distribution (OOD) detection and the emerging field of adversarially robust OOD detection.

Post-hoc OOD detectors are designed to function with pre-trained models, making them highly practical for real-world deployment. 
These methods can be broadly categorised based on the model information they utilise.

The earliest and simplest methods operate directly on the model's final output layer. 
The Maximum Softmax Probability (MSP)~\cite{hendrycks_baseline_2017} uses the model's confidence score as a direct indicator of in-distribution behaviour. 
Subsequent work proposed the energy score~\cite{liu_energy-based_2020}, which offers a more theoretically grounded and often more effective scoring function derived from the logit values. 
GEN~\cite{liu_gen_2023} computes a generalised entropy score to further amplify the separation of ID and OOD inputs. 
While computationally inexpensive, these methods are fundamentally limited by their reliance on the output layer and are known to be vulnerable to the overconfidence issue in DNNs~\cite{nguyen_deep_2015}, where a model can produce a high-confidence prediction for a novel input.

 To overcome the limitations of output-based scores, a significant body of research has turned to the model's intermediate feature representations, which provide a richer signal for detecting novelty.

Activation-based methods manipulate or analyse neuron activation patterns. 
ReAct~\cite{sun_react_2021} improves OOD detection by clipping high-magnitude activations, which are found to be more prevalent in OOD inputs. 
More recent techniques, such as ASH~\cite{djurisic2022extremely}, propose even more advanced activation shaping to further separate ID and OOD representations.

Distance-based methods compute a score based on the distance of a test input's feature embedding to the learned distribution of in-distribution data~\cite{stoica2025outofdistributiondetectionusingcounterfactual}. Prominent examples include using the Mahalanobis distance~\cite{lee_simple_2018} or the distance to the k-nearest neighbour (KNN)~\cite{sun_out--distribution_2022} in the feature space. 
The Fast Decision Boundary based Detector (fDBD)~\cite{liu_fast_2024}, approximates an input's distance to the learned class decision boundaries, operating on the principle that OOD points tend to reside closer to these boundaries than ID points.

Perturbation-based scores have also been explored to distinguish between ID and OOD data. 
ODIN~\cite{dcbe7abf4db64d1b89bf9802585660ed} applies small, gradient-based perturbations to input data that intentionally increase softmax confidence, as well as applying temperature scaling to the softmax output. Since ID samples tend to gain more trust from these perturbations than OOD samples do, this results in a more precise separation in detection scores.
PRO~\cite{chen_leveraging_2025} applies multi‑step adversarial-style perturbations to input data and uses the resulting confidence drop, which is more pronounced for OOD inputs than in‑distribution ones, as an enhanced detection signal.

Despite their increasing sophistication, many OOD detectors remain vulnerable to adversarial attacks. 
An adversarial attack in this context aims to generate an OOD input that the detector misclassifies as in-distribution, effectively creating a ``blind spot" in the system's safety monitor.

Recent work has systematically demonstrated the vulnerability of even state-of-the-art detectors to adversarial attacks.~\cite{azizmalayeri2022your,lorenz_deciphering_2024}. 
In response to these threats, research into robust OOD detection is emerging. 
One defence strategy is adversarial training, where the model is retrained on a dataset augmented with adversarial examples.
This method is used by a few OOD detectors such as ALOE~\cite{chen_robust_2021} and ATOM~\cite{chen_atom_2021}.
While this can improve robustness, it is computationally expensive and sacrifices the primary benefit of post-hoc methods: applicability to already-trained models.

In contrast to other robust OOD detection methods, ROSS is entirely post-hoc and applicable to pre-trained DNNs.
Additionally, compared to PRO, ROSS can quantify the stability of the OOD score landscape rather than simply decreasing the DNN's confidence, thereby creating a more robust and reliable OOD detector.
\section{Preliminaries}\label{sec:preliminaries}
We begin by defining the notation that will be used throughout this work and providing the necessary theoretical background.

\textbf{Neural Networks.}
Our framework is based on a standard classification setting. 
The inputs are drawn from a $d$-dimensional feature space $\mathcal{X} \subseteq \mathbb{R}^d$, while the outputs belong to a set $\mathcal{Y} = \{1, 2, \dots, C\}$ of $C$ possible classes~\cite{goodfellow_deep_2016}. 
We are given a training set, $\mathcal{D}_{\text{train}} = \{(x^{(i)}, y^{(i)})\}_{i=1}^{m}$, consisting of $m$ such input-output pairs. 
The classifier is a deep neural network (DNN), $f$, trained on $\mathcal{D}_{\text{train}}$. 
We specifically denote the feature representation from the penultimate layer of this network as $f_{-1}(x)$.

\textbf{OOD Detection.}
Once a DNN is deployed and used in the real world, it may encounter inputs unlike those it was trained and tested on. 
These types of inputs are considered OOD.
The goal of DNN OOD detection is to identify the outputs for these points as unsafe predictions, which may trigger a warning system or necessitate human intervention.
This is done by creating a score function $S: \mathcal{X} \rightarrow \mathbb{R}$ whose output is used to determine whether an input is OOD or not by some threshold $\tau$.
Specifically, if $S(x) \geq \tau$, an input is classified as ID and otherwise OOD. 

OpenOOD~\cite{yang_openood_2022} defines two subsets of OOD data: near-OOD and far-OOD.
Near-OOD datasets are primarily characterised by a semantic shift from the in-distribution data, whereas far-OOD datasets additionally exhibit a distinct shift in their domain or covariate features.
Far-OOD points are typically easier for a model to identify and assign a low confidence score.

\textbf{Robust OOD Detection.}
Reliable OOD detectors should maintain sound OOD separation even under adversarial conditions.
Specifically, they should handle inputs $x$ where small perturbations \( \delta \) have been added to alter the OOD score $S(x)$ to fool the detector.
There are two ways to attack an OOD detector: by creating adversarial \emph{ID} inputs or creating adversarial \emph{OOD} inputs.

More formally, we can create an adversarial ID point by
\begin{equation}
x_{\text{adv}} = x + \delta, \quad \text{where} \quad \delta = \arg\min_{\|\delta\| \leq \epsilon} S(x + \delta),
\end{equation}
where \( S(x) \) is a chosen OOD scoring function, such as MSP, energy-based scoring, or Mahalanobis distance.
We minimise $S(x)$ to identify the ID point as OOD.
Similarly, to create an adversarial OOD point, we maximise $S(x)$ to create an OOD point that is classed as ID.

Achieving robust OOD detection is a critical step toward deploying machine learning systems that operate reliably in open-world environments, where unknown or malicious inputs are inevitable.

\section{ROSS}\label{sec:methodology}

\begin{figure*}[t!]
\centering
\hspace{0.4cm}
\makebox[0.14\textwidth]{\centering\small\textbf{CIFAR-100}}%
\hfill
\makebox[0.14\textwidth]{\centering\small\textbf{TIN}}%
\hfill
\makebox[0.14\textwidth]{\centering\small\textbf{MNIST}}%
\hfill
\makebox[0.14\textwidth]{\centering\small\textbf{SVHN}}%
\hfill
\makebox[0.14\textwidth]{\centering\small\textbf{Texture}}%
\hfill
\makebox[0.14\textwidth]{\centering\small\textbf{Places365}}

\subfloat{
    \raisebox{0.1cm}{\rotatebox{90}{\small\textbf{Median ($\smed$)}}}
}
\hspace{0.1cm}
\subfloat{\includegraphics[width=0.16\textwidth]{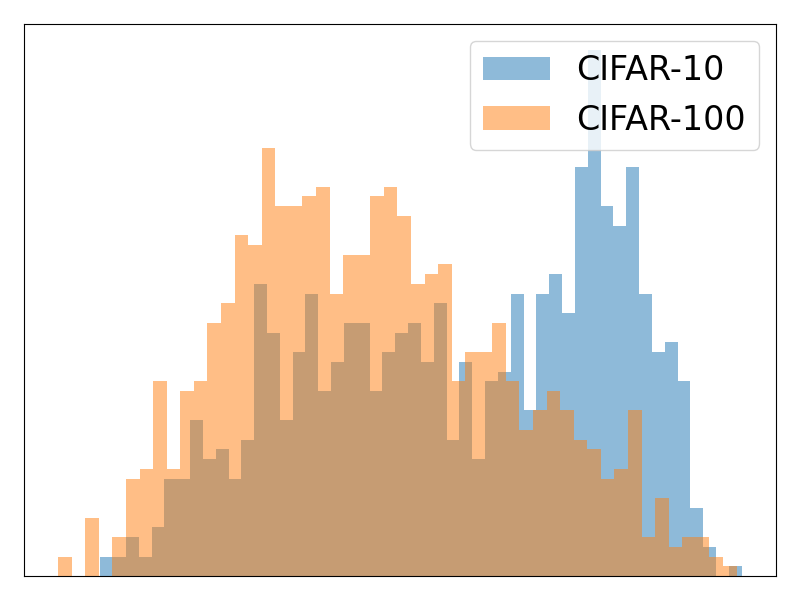}\label{fig:hist_a}}
\hfill
\subfloat{\includegraphics[width=0.16\textwidth]{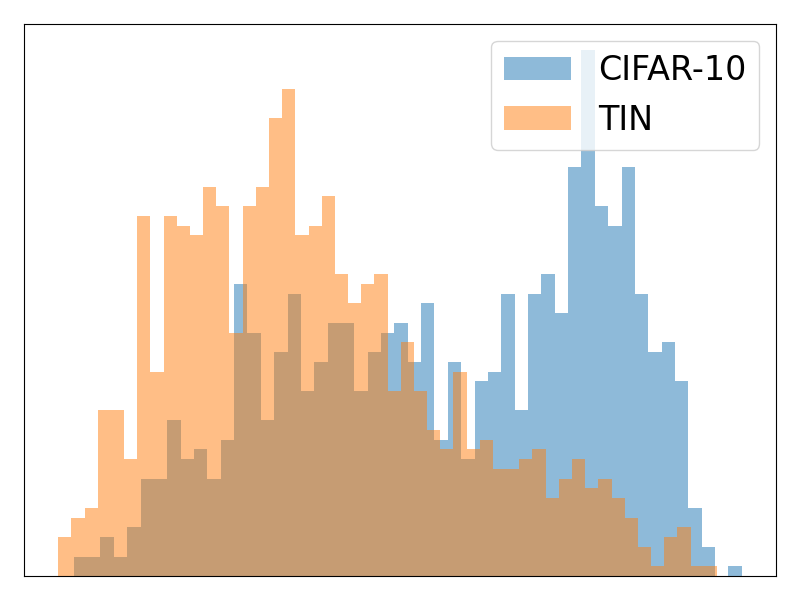}\label{fig:hist_b}}
\hfill
\subfloat{\includegraphics[width=0.16\textwidth]{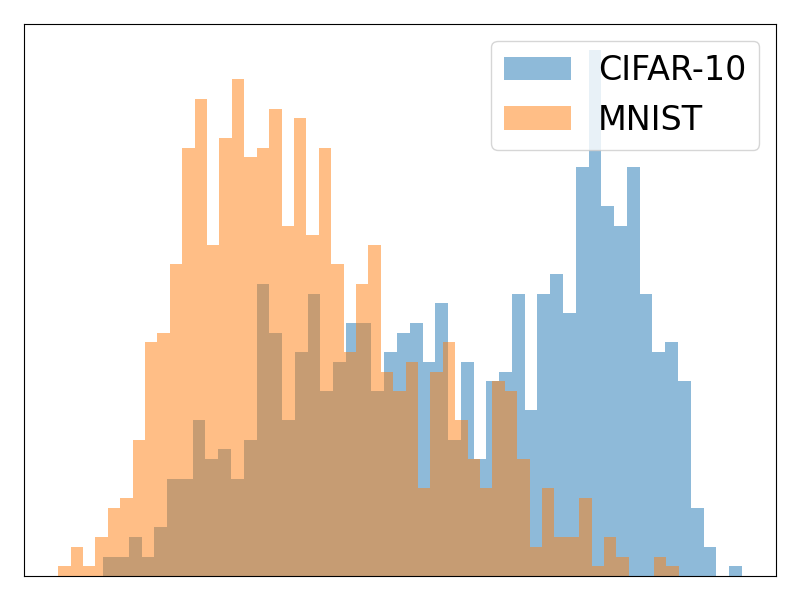}\label{fig:hist_c}}
\hfill
\subfloat{\includegraphics[width=0.16\textwidth]{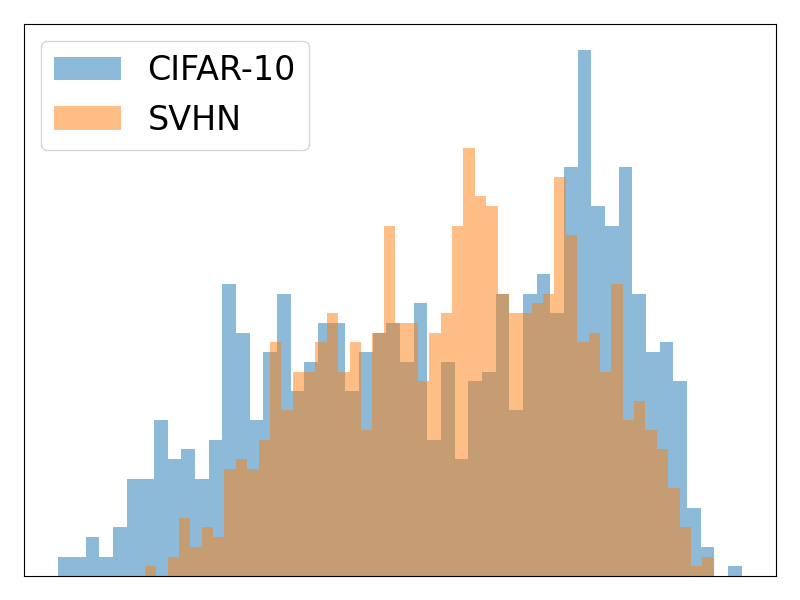}\label{fig:hist_d}}
\hfill
\subfloat{\includegraphics[width=0.16\textwidth]{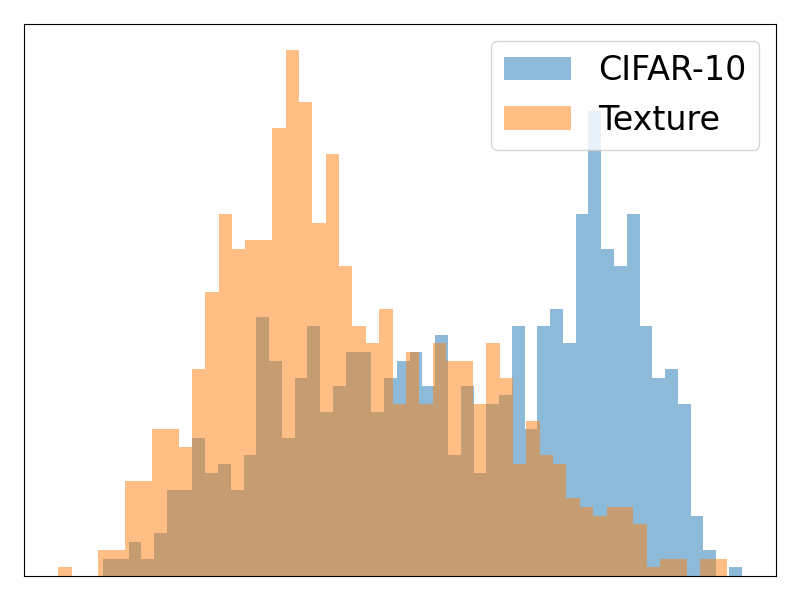}\label{fig:hist_e}}
\hfill
\subfloat{\includegraphics[width=0.16\textwidth]{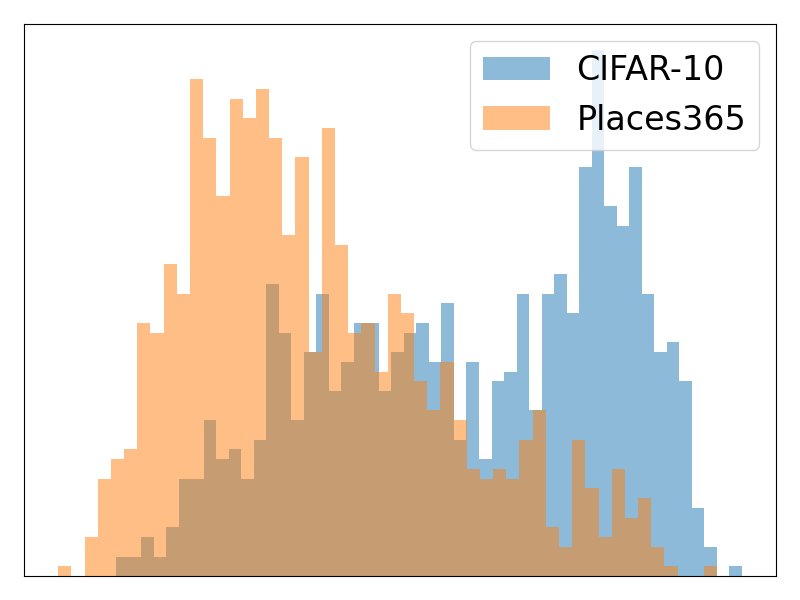}\label{fig:hist_f}}

\subfloat{
    \raisebox{0.1cm}{\rotatebox{90}{\small\textbf{(-) MAD ($\mad$)}}}
}
\hspace{0.1cm}
\subfloat{\includegraphics[width=0.16\textwidth]{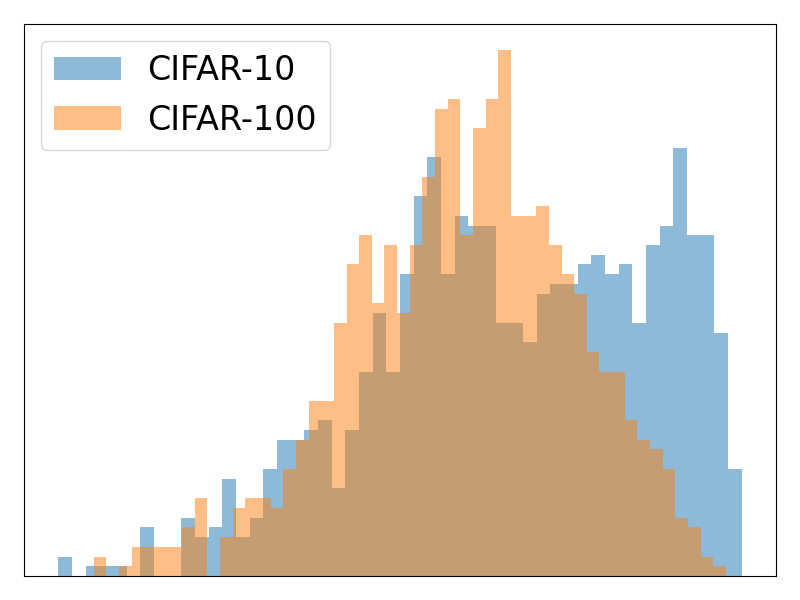}\label{fig:hist_g}}
\hfill
\subfloat{\includegraphics[width=0.16\textwidth]{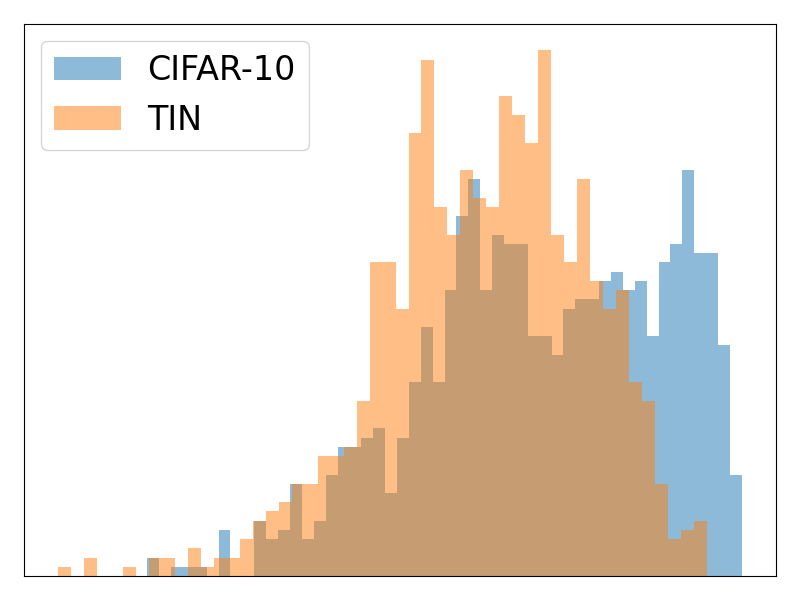}\label{fig:hist_h}}
\hfill
\subfloat{\includegraphics[width=0.16\textwidth]{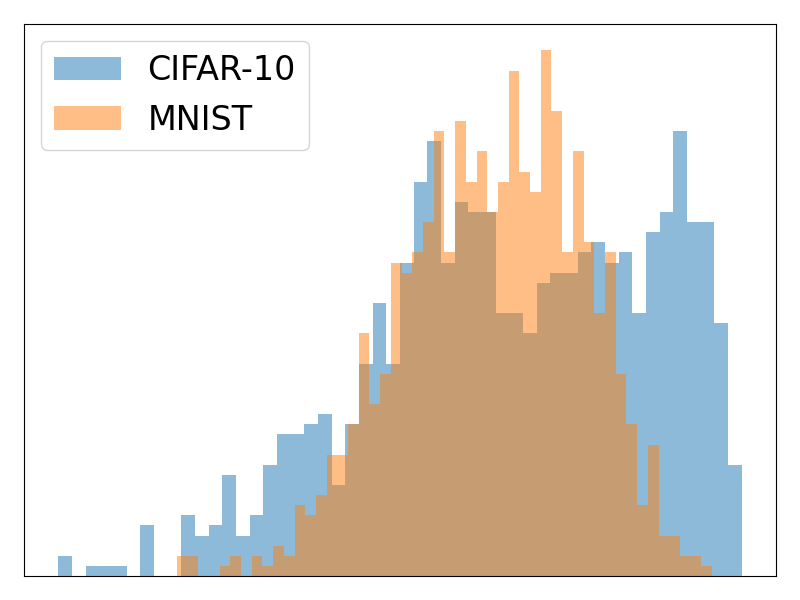}\label{fig:hist_i}}
\hfill
\subfloat{\includegraphics[width=0.16\textwidth]{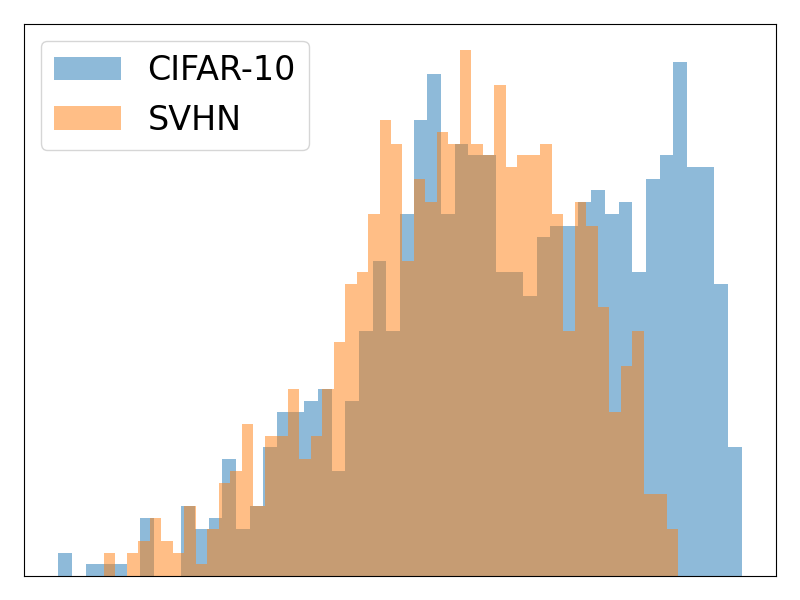}\label{fig:hist_j}}
\hfill
\subfloat{\includegraphics[width=0.16\textwidth]{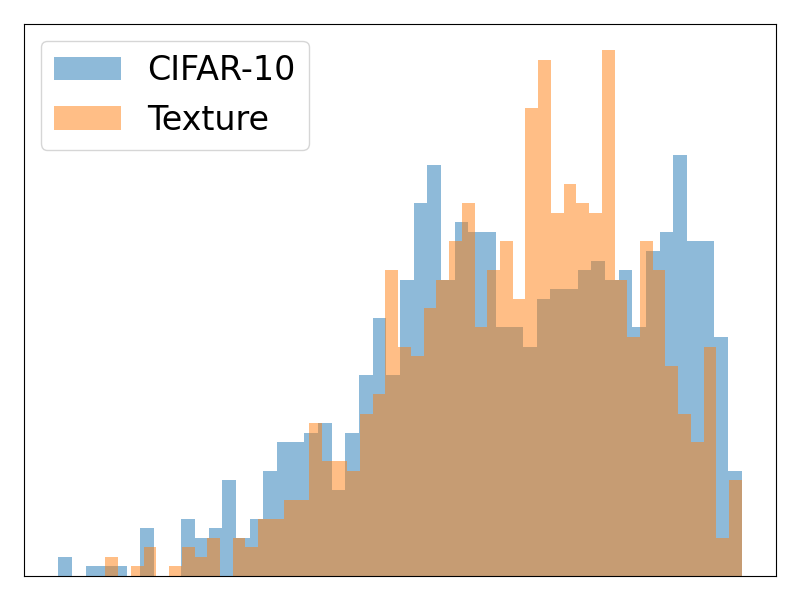}\label{fig:hist_k}}
\hfill
\subfloat{\includegraphics[width=0.16\textwidth]{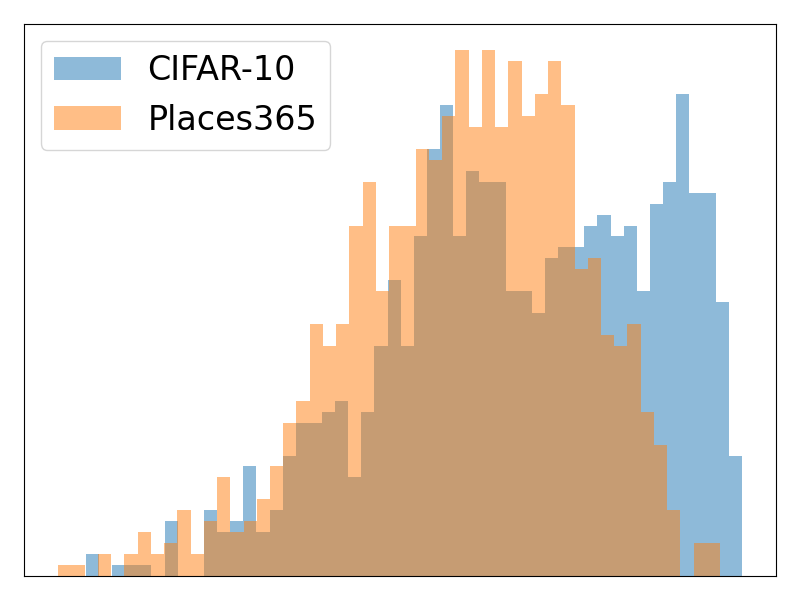}\label{fig:hist_l}}

\subfloat{
    \raisebox{0.6cm}{\rotatebox{90}{\small\textbf{ROSS}}}
}
\hspace{0.1cm}
\subfloat{\includegraphics[width=0.16\textwidth]{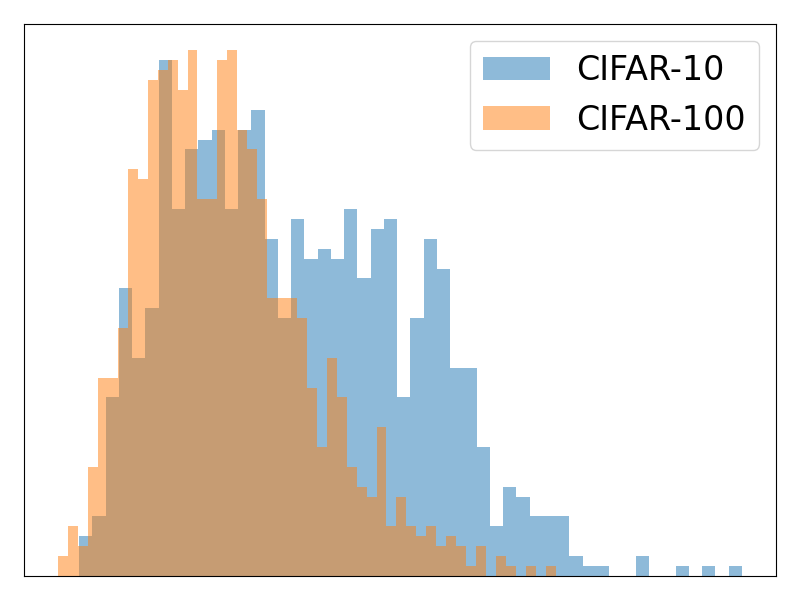}\label{fig:hist_m}}
\hfill
\subfloat{\includegraphics[width=0.16\textwidth]{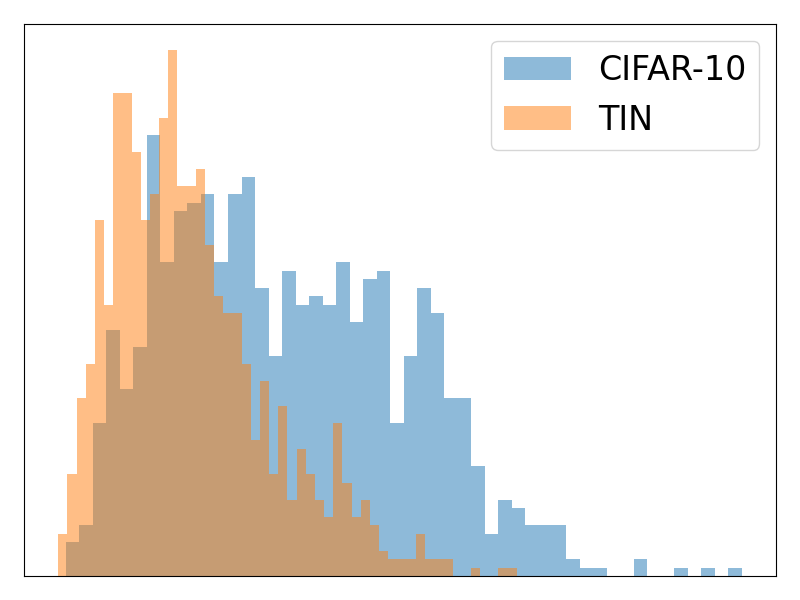}\label{fig:hist_n}}
\hfill
\subfloat{\includegraphics[width=0.16\textwidth]{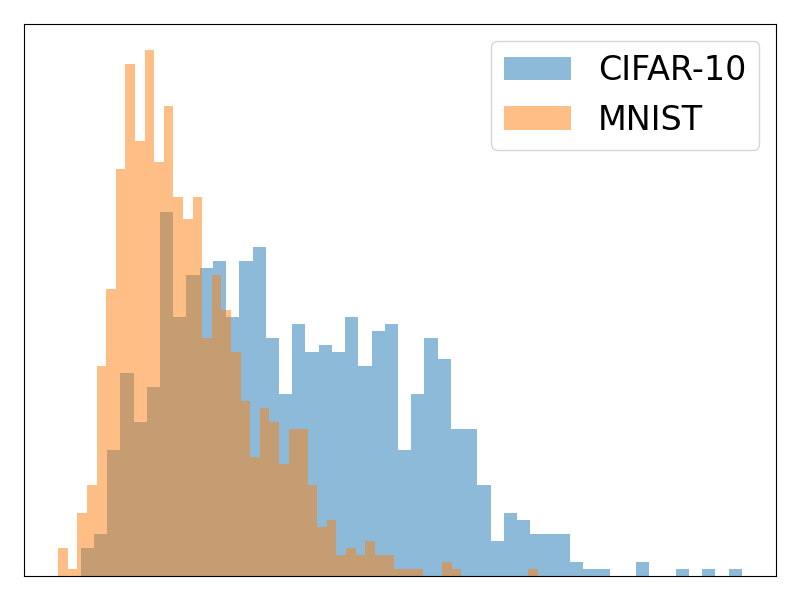}\label{fig:hist_o}}
\hfill
\subfloat{\includegraphics[width=0.16\textwidth]{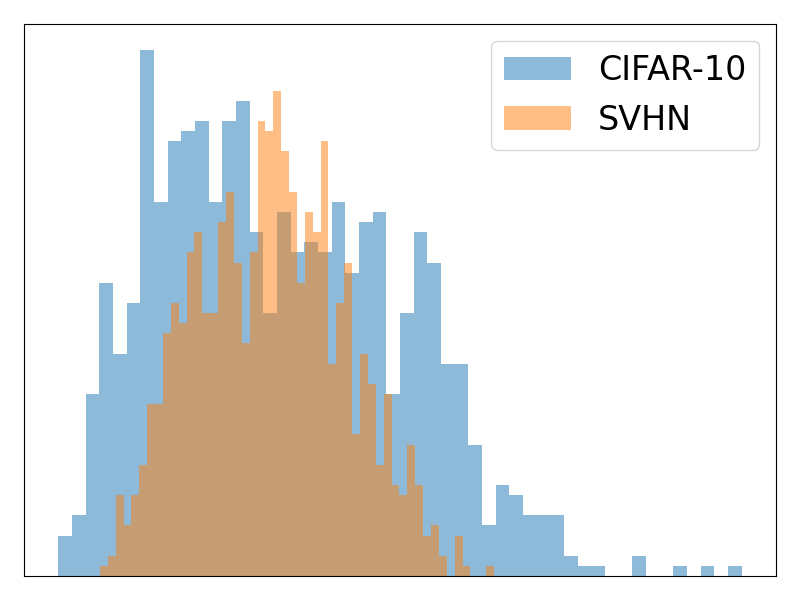}\label{fig:hist_p}}
\hfill
\subfloat{\includegraphics[width=0.16\textwidth]{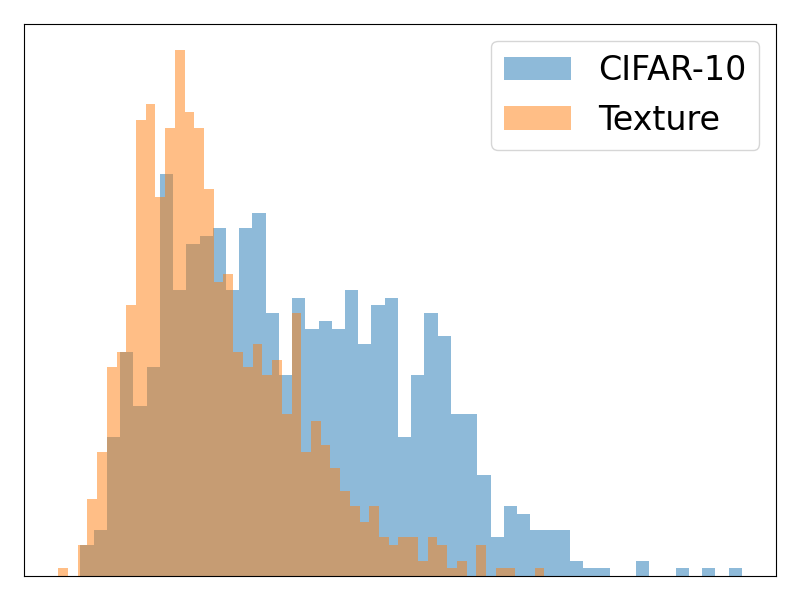}\label{fig:hist_q}}
\hfill
\subfloat{\includegraphics[width=0.16\textwidth]{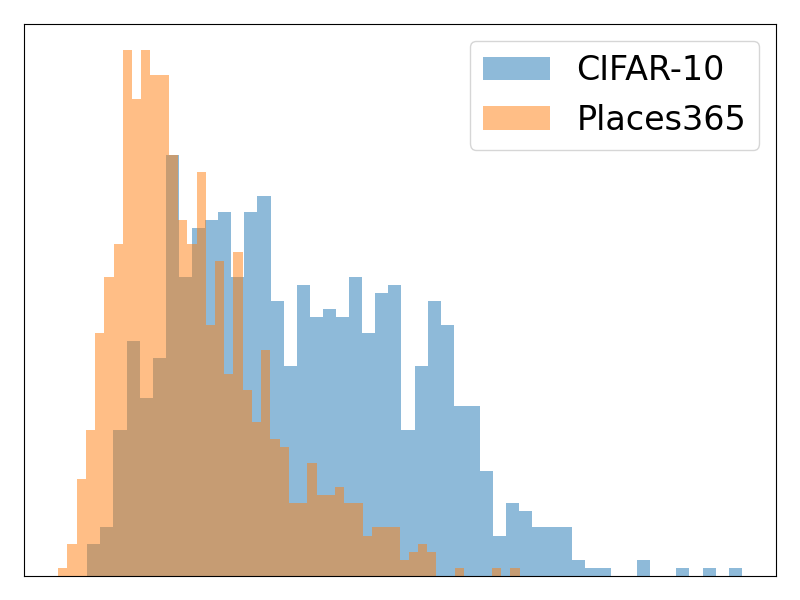}\label{fig:hist_r}}

\caption{
\textbf{Distributions of Median, MAD, and ROSS scores for ID (CIFAR-10) and various OOD datasets (e.g., SVHN, Texture).} Columns correspond to different OOD datasets; rows show Median Smoothed Score (top), MAD (middle, negated so ID is larger), and ROSS (bottom). Blue: ID, orange: OOD. ROSS yields the best ID–OOD separation.
}
\label{fig:score_histograms}
\end{figure*}

In this section, we introduce ROSS, a novel, perturbation-based scoring method for OOD detection. 
Our approach is designed to be robust against adversarial attacks by leveraging median smoothing and the instability of the model's score landscape.

\subsection{Perturbation-Based Scoring}
Given an input $x$, ROSS generates a set of $N$ noisy samples $\{{x}'_1, {x}'_2, \dots, {x}'_N\}$ by adding small random perturbations.  
Each noisy sample ${x}'_i$ is passed through the base OOD scoring function to produce a collection of scores, forming the \textbf{score stack}.  
From this stack, we derive two robust statistics:
\begin{enumerate}
    \item \textbf{Median Score ($\smed$):} the median of the score stack, representing the smoothed, central tendency of the OOD score.
    \item \textbf{Score Stability ($\mad$):} the Median Absolute Deviation (MAD) of the score stack, capturing the local instability of the score landscape.
\end{enumerate}

These two quantities jointly summarise the model’s confidence and its local sensitivity to perturbations.  
Because their computation is non-directional and gradient-free, they are inherently resistant to gradient-based adversarial manipulation.  
As a result, ROSS achieves symmetric robustness, maintaining consistent performance under both score-minimising (PGD-min) and score-maximising (PGD-max) attacks.  

Our core hypothesis is that in-distribution (ID) inputs exhibit smoother, more stable score landscapes (low $\mad$), whereas out-of-distribution (OOD) inputs produce higher variance under perturbations.
To test this theory, we analyse the results using the base GEN~\cite{liu_gen_2023} score, following the methodology of PRO~\cite{chen_leveraging_2025}.
Figure~\ref{fig:score_histograms} illustrates how $\smed$ and $\mad$ individually separate ID and OOD samples, but ROSS provides better discrimination, showing that a naive combination of these signals is insufficient for robust detection.
In Section~\ref{sec:evaluation}, we provide results to demonstrate this further.

\begin{algorithm}[tb]
\caption{ROSS OOD Scoring}
\label{alg:ross}
\textbf{Input}: Input sample $x$, base scoring function $\text{BaseScore}(\cdot)$ \\
\textbf{Parameter}: Number of perturbations $N$, noise magnitude $\sigma_{\text{noise}}$, stability hyperparameter $\lambda$, confidence threshold $\st$. \\
\textbf{Output}: Final ROSS score $\sclam$.\\
\begin{algorithmic}[1]
\STATE $\mathcal{S} \leftarrow []$ \hfill{\COMMENT{Initialise empty score list.}}
\FOR{$i=1$ \TO $N$}
    \STATE $\epsilon \sim \mathcal{N}(0, \sigma_{\text{noise}}^2 I)$ \hfill{\COMMENT{Generate perturbation.}}
    \STATE  ${x}' \leftarrow x + \epsilon$ \hfill{\COMMENT{Create noisy sample.}} 
    \STATE $s \leftarrow \text{BaseScore}({x}')$.\hfill{\COMMENT{Calculate base score.}} 
    \STATE Append $s$ to $\mathcal{S}$.
\ENDFOR
\STATE
\STATE $\smed \leftarrow \text{Median}(\mathcal{S})$ \hfill{\COMMENT{Calculate median score.}}
\STATE $\mad \leftarrow \text{MAD}(\mathcal{S})$  \hfill{\COMMENT{Calculate MAD of scores.}}
\STATE
\STATE $\Delta_{\text{score}} \leftarrow \max(0, \smed - \st)$ \hfill{\COMMENT{Calculate gated score.}}
\STATE $\sclam \leftarrow \min(\st,\smed) + \Delta_{\text{score}} \cdot \left(1+\frac{\lambda}{\mad}\right)$
\STATE
\RETURN $\sclam$
\end{algorithmic}
\end{algorithm}

\subsection{Confidence-Thresholded Scoring}
A challenge arises because some ``far-OOD'' samples may produce highly stable but consistently low-confidence scores.  
Naively rewarding stability would incorrectly enhance these samples, thereby reducing their discriminative power.  

To mitigate this, we introduce a confidence-thresholded scoring function that rewards stability only for samples already deemed plausible ID candidates.  
Specifically, the stability bonus is applied only when the median score $\smed$ exceeds a confidence threshold $\st$, defined as the 5th percentile of the $\smed$ distribution on a held-out ID validation set.

The final score is:
\begin{equation} \label{eq:clamped_score}
    \sclam = \min(\st, \smed) + \Delta_{\text{score}} \cdot \left(1+\frac{\lambda}{\mad}\right),
\end{equation}
where:
\begin{itemize}
    \item $\st$ is the 5th percentile confidence threshold.
    \item $\mad$ is the Median Absolute Deviation of the score stack.
    \item $\Delta_{\text{score}} = \max(0, \smed - \st)$ ensures the stability bonus applies only when $\smed > \st$.
    \item $\lambda$ controls the magnitude of the stability bonus.
\end{itemize}

This formulation ensures that only high-confidence, locally stable samples receive a score boost.  
The magnitude of this boost is inversely proportional to $\mad$, clamping confident, stable samples to the upper end of the score distribution and improving separation from OOD inputs.

\subsection{Theoretical Motivation}
ROSS is theoretically grounded in robust statistics, leveraging estimators known to resist outliers and adversarial corruption~\cite{Hampel01061974}.  
Let $\text{BaseScore}(x)$ denote an OOD scoring function that assigns higher values to ID samples and lower values to OOD samples.  
For ID inputs, assuming local smoothness of $\text{BaseScore}(\cdot)$, adding Gaussian noise $\epsilon \sim \mathcal{N}(0, \sigma^2 I)$ yields perturbed samples ${x}' = x + \epsilon$ whose scores cluster tightly around $\text{BaseScore}(x)$.  
Thus, their local score variance remains low.  
Conversely, OOD inputs typically inhabit irregular regions of the score surface, where small perturbations induce large fluctuations.

Our method quantifies this instability through the median absolute deviation, $\mad$, and penalises high instability via the inverse-MAD term in Eq.~\ref{eq:clamped_score}.  
However, rewarding stability across all inputs risks inflating far-OOD scores that happen to be uniformly low.  
The gating term $\max(0, \smed - \st)$ prevents this by ensuring that only ID-like inputs benefit from the stability bonus.  
Using the median instead of the mean further reduces sensitivity to extreme scores introduced by adversarial perturbations~\cite{chiang_detection_2020}.

This mechanism naturally enforces two desirable properties:
\begin{itemize}
    \item \textbf{Stability selectivity:} Only ID-like regions with smooth, stable score behaviour benefit from the stability bonus.
    \item \textbf{Adversarial robustness:} The median and MAD provide empirical resistance to gradient-based perturbations and score outliers~\cite{Cohen19}.
\end{itemize}

Together, these components yield a scoring function that adaptively rewards both high confidence and local score stability, two key hallmarks of in-distribution behaviour.  
This principled formulation enables ROSS to maintain strong detection accuracy under both natural and adversarial distribution shifts.
\begin{figure}
    \centering
    \includegraphics[width=0.95\linewidth, 
    ]
    {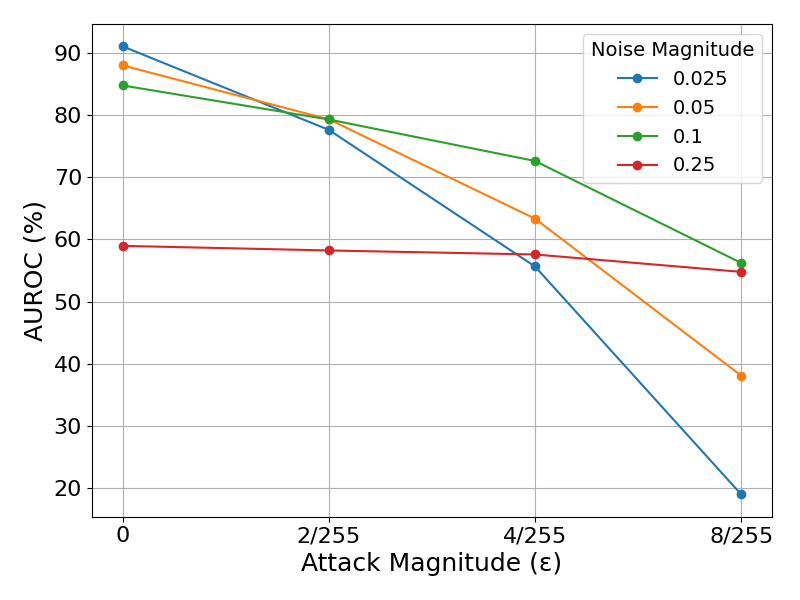}
    \caption{\textbf{Trade-off between clean and adversarial (PGD-Max) performance.} 
     OOD detection AUROC on CIFAR-10 under attacks of varying strength ($\epsilon$). Each line shows a different ROSS noise level ($\sigma_{\text{noise}}$), revealing the balance between clean accuracy ($\epsilon=0$) and robustness.
    }
    \label{fig:noise-tradeoff}
\end{figure}
\section{Evaluation}\label{sec:evaluation}
 \begin{table*}[ht!]
 \centering
 \caption{
Analysis of proposed post-processors using the base GEN score. The model is trained on CIFAR-10 (ID) and evaluated against various OOD benchmarks. Performance is reported as FPR95 (\%) $\downarrow$ / AUROC (\%) $\uparrow$. Best metric is in \textbf{bold} and second best is \underline{underlined}.
 }
 \label{tab:ross_gen_cifar10_comparison}
 \begin{tabular}{lcccccc|c}
 \toprule
 \textbf{Post-processor} & \multicolumn{2}{c}{\textbf{near-OOD}} & \multicolumn{4}{c|}{\textbf{far-OOD}} & \textbf{Avg.} \\
 \cmidrule(lr){2-3} \cmidrule(lr){4-7}
 & \textbf{CIFAR-100} & \textbf{TIN} & \textbf{MNIST} & \textbf{SVHN} & \textbf{Texture} & \textbf{Places365} & \\
 \midrule
 $\smed$ & 61.67/\underline{82.95} & \underline{52.66}/\underline{85.59} & \underline{51.39}/\underline{84.47} & 59.82/76.30 & \underline{53.75}/\textbf{84.66} & \underline{43.43}/\underline{88.38} & \underline{53.79}/\underline{83.73} \\
 $\mad$ & \underline{58.65}/72.68 & 56.27/73.29 & 61.14/64.69 & \underline{47.67}/\underline{78.22} & 84.19/61.08 & 55.94/72.73 & 60.64/70.45 \\
 $\smed/\mad$ & 96.19/36.16 & 95.67/36.87 & 85.17/46.24 & 97.86/25.86 & 91.59/49.73 & 94.76/39.25 & 93.54/39.02 \\
 $\sclam$ & \textbf{49.90}/\textbf{84.09} & \textbf{44.26}/\textbf{86.53} & \textbf{44.28}/\textbf{85.23} & \textbf{46.89}/\textbf{78.80} & \textbf{50.33}/\underline{84.59} & \textbf{38.98}/\textbf{89.05} & \textbf{45.77}/\textbf{84.72} \\
 \bottomrule
 \end{tabular}
 \end{table*}
We empirically evaluate the effectiveness of ROSS across standard OOD benchmarks. 
We begin by validating the design of ROSS through comparisons with its components, namely, the median-smoothed score and the score stability metric, to demonstrate how ROSS effectively integrates both aspects into a more balanced and discriminative score. 
We then assess the robustness of ROSS under adversarial conditions by benchmarking it against established OOD detection methods subjected to gradient-based attacks.

\subsection{Experimental Setup}
\textbf{Datasets.}
We evaluate our method using three ID datasets: CIFAR-10, CIFAR-100~\cite{krizhevsky_learning_2009} and ImageNet~\cite {deng_imagenet_2009}. To ensure comparability with existing benchmarks, we adopt the standardised evaluation protocol from OpenOOD~\cite{zhang_openood_2024}.
OOD datasets are categorised as either near-OOD, which are semantically different but visually similar to the ID data, or far-OOD, which differ both visually and semantically and lie further from the ID distribution.

For CIFAR-10, near-OOD datasets are CIFAR-100 and TinyImageNet (TIN)~\cite{le2015tiny}, and far-OOD datasets are MNIST~\cite{lecun_mnist_1998}, SVHN~\cite{netzer_reading_2011}, Texture~\cite{cimpoi_describing_2014}, and Places365~\cite{zhou_places_2018}.
For CIFAR-100, we use CIFAR-10/TIN as near-OOD and the same far-OOD sets.
ImageNet results (Appendix Table~\ref{tab:imagenet_ross_gen}) follow the OpenOOD setup.

\textbf{Networks.}
We use pre-trained ResNet-18~\cite{he_deep_2016} networks for CIFAR-10 and CIFAR-100, and a pre-trained ResNet-50 network from OpenOOD~\cite{zhang_openood_2024} for ImageNet.
Results are produced by averaging over the three training runs provided.

\textbf{Adversarial Robustness Evaluation.}
To evaluate the robustness of OOD detection methods under adversarial conditions, we apply two types of white-box attacks adapted for OOD detection: \textbf{PGD-min} and \textbf{PGD-max}, following the setup in prior work~\cite{chen_atom_2021, kopetzki_evaluating_2021}.
These attacks target the input by minimising the ID score or maximising the OOD score, respectively, to induce incorrect OOD classifications.

We use the Projected Gradient Descent (PGD) method with $\ell_\infty$ perturbations constrained by a radius $\epsilon$~\cite{madry2018towards}. 
For each evaluation, we vary $\epsilon$ across a range of values to assess robustness under increasing attack strength. 
Attacks are run for 40 steps, and gradients are computed with respect to the base OOD scoring function being evaluated.
 
\textbf{Hyperparameters.}
Our method introduces three primary hyperparameters: the number of perturbed samples \( N \), the perturbation noise magnitude \( \sigma_{\text{noise}} \), and the stability bonus weight \( \lambda \) (as defined in Section~\ref{sec:methodology}).

We found that setting \( N = 25 \) achieves a strong balance between computational efficiency and robustness. While using fewer samples can lead to less stable estimates of the median and MAD statistics, increasing \( N \) beyond 25 yields diminishing returns in terms of detection accuracy.
However, performance remains strong with $N=5$ or $N=10$, and a trade-off between robustness and computational cost needs to be balanced.
We provide an ablation study in the appendix in Table~\ref{tab:table_n_ablation_study}.

The magnitude of the noise controls the trade-off between sensitivity and robustness. 
Smaller values may fail to expose instability in the score landscape, while larger noise can excessively distort the input. 
Through empirical validation, we set $\sigma_{\text{noise}} = 0.1$, which strikes a balance between standard and clean accuracy.
We present a comparison in Figure~\ref{fig:noise-tradeoff} where increasing noise reduces clean accuracy but increases robust accuracy.
We also present an ablation study showing this trade-off in the appendix, Table~\ref{tab:robustness_vs_noise}.

A higher $\lambda$ amplifies the reward for score stability among high-confidence inputs. We selected $\lambda = 0.05$ based on validation performance.
We provide ablation studies in the appendix for CIFAR-10 (Table~\ref{tab:ross_lambda_variants}) and CIFAR-100 (Table~\ref{tab:ross_lambda_variants_cifar100}).

\textbf{Hardware.}
All experiments were conducted on a machine equipped with an Intel Core i9-10940X 14-core CPU, 256 GB of RAM, and an NVIDIA RTX 4060 Ti GPU with 16 GB of VRAM.

\subsection{Evaluation of Post-Processors}
\begin{table*}[ht!]
\centering
\caption{
Robustness of OOD detection scores for CIFAR-10 against PGD-\textbf{min} and PGD-\textbf{max} attacks with varying attack radii ($\epsilon$).
Results are averaged across all benchmarks, and performance is reported as FPR95 (\%) $\downarrow$ / AUROC (\%) $\uparrow$. Our proposed method is highlighted in grey. Best metric is in \textbf{bold} and second best is \underline{underlined}.
}
\label{tab:combined_attacks_cifar10}
\resizebox{\textwidth}{!}{%
\begin{tabular}{l|cc|cc|cc}
\toprule
\textbf{Post-processor} & \multicolumn{2}{c|}{\textbf{$\epsilon=2/255$}} & \multicolumn{2}{c|}{\textbf{$\epsilon=4/255$}} & \multicolumn{2}{c}{\textbf{$\epsilon=8/255$}} \\
\cmidrule(lr){2-3} \cmidrule(lr){4-5} \cmidrule(lr){6-7}
& \textbf{PGD-Min} & \textbf{PGD-Max} & \textbf{PGD-Min} & \textbf{PGD-Max} & \textbf{PGD-Min} & \textbf{PGD-Max} \\
\midrule
MSP & 74.24/68.44 & 92.59/57.81 & 92.40/48.73 & 99.27/25.06 & 99.53/27.89 & 99.99/3.76 \\
EBO & 83.21/72.34 & 93.80/55.20 & 95.94/51.22 & 99.34/22.85 & 99.82/20.79 & 99.99/3.24 \\
GEN & 77.84/72.01 & 93.09/56.93 & 93.97/52.36 & 99.31/24.07 & 99.70/26.04 & 99.99/3.51 \\
ODIN & 89.03/69.08 & 93.22/52.57 & 97.56/50.43 & 98.84/25.85 & 99.92/24.81 & 99.98/5.62 \\
fDBD & 59.92/76.23 & 78.99/69.51 & 82.87/58.89 & 96.00/43.52 & 97.25/35.54 & 99.64/17.86 \\
PRO-GEN & 58.60/81.80 & 79.16/73.01 & 78.11/71.85 & 96.73/41.28 & 94.08/55.99 & 99.82/10.54 \\
PRO-fDBD & \underline{54.07}/\textbf{82.65} & 61.83/74.88 & 73.79/\underline{72.38} & 93.38/47.59 & 92.13/54.11 & 99.50/19.62 \\
\midrule
\rowcolor{gray!15} ROSS-MSP & 56.09/77.65 & \underline{55.87}/77.65 & \underline{68.32}/69.71 & \underline{69.03}/69.98 & \underline{85.99}/52.29 & 89.83/52.65 \\
\rowcolor{gray!15} ROSS-EBO & 63.18/78.01 & 59.94/\underline{80.57} & 75.06/71.48 & 79.87/69.96 & 90.74/\underline{56.99} & 93.67/51.17 \\
\rowcolor{gray!15} ROSS-GEN & 58.42/78.75 & 58.80/78.78 & 70.83/71.95 & 70.35/\underline{71.62} & 88.30/56.85 & \underline{88.08}/\underline{54.32} \\
\rowcolor{gray!15} ROSS-fDBD & \textbf{52.46}/\underline{80.98} & \textbf{51.05}/\textbf{81.02} & \textbf{64.47}/\textbf{74.34} & \textbf{60.51}/\textbf{74.58} & \textbf{84.39}/\textbf{59.10} & \textbf{81.99}/\textbf{58.81} \\
\bottomrule
\end{tabular}
}
\end{table*}

\begin{table*}[ht!]
\centering
\caption{
Robustness of OOD detection scores for CIFAR-100 against PGD-\textbf{min} and PGD-\textbf{max} attacks with varying attack radii ($\epsilon$). Results are averaged across all benchmarks, and performance is reported as FPR95 (\%) $\downarrow$ / AUROC (\%) $\uparrow$. Our proposed method is highlighted in grey. Best metric is in \textbf{bold} and second best is \underline{underlined}.
}
\label{tab:combined_attacks_cifar100}
\resizebox{\textwidth}{!}{%
\begin{tabular}{l|cc|cc|cc}
\toprule
\textbf{Post-processor} & \multicolumn{2}{c|}{\textbf{$\epsilon=2/255$}} & \multicolumn{2}{c|}{\textbf{$\epsilon=4/255$}} & \multicolumn{2}{c}{\textbf{$\epsilon=8/255$}} \\
& \textbf{PGD-Min} & \textbf{PGD-Max} & \textbf{PGD-Min} & \textbf{PGD-Max} & \textbf{PGD-Min} & \textbf{PGD-Max} \\
\midrule
MSP & 86.75/49.66 & 90.33/43.06 & 97.30/32.40 & 98.90/21.83 & 99.81/18.46 & 99.97/5.73 \\
EBO & 88.27/48.83 & 91.77/47.35 & 98.17/20.47 & 99.12/22.73 & 99.95/2.11 & 99.98/5.23 \\
GEN & 86.72/49.92 & 90.58/47.02 & 97.49/23.24 & 98.87/23.29 & 99.71/3.69 & 99.88/5.87 \\
ODIN & 88.37/54.01 & 88.98/52.81 & 98.49/27.76 & 97.53/31.36 & 99.99/4.75 & 99.76/10.90 \\
fDBD & 83.11/54.74 & 84.29/51.21 & 95.53/32.94 & 96.54/30.40 & 99.65/11.03 & 99.72/11.02 \\
PRO-GEN & 84.91/50.33 & 87.33/51.45 & 96.06/24.03 & 98.74/22.78 & 99.56/3.71 & 99.93/3.93 \\
PRO-fDBD & 81.35/58.93 & 83.72/50.76 & 94.01/38.82 & 96.46/28.38 & 99.29/15.18 & 99.76/9.31 \\
\midrule
\rowcolor{gray!15} ROSS-MSP & 79.08/60.73 & 77.61/61.12 & 85.01/52.73 & 82.57/52.84 & 93.57/38.00 & 90.30/37.54 \\
\rowcolor{gray!15} ROSS-EBO & 77.20/62.96 & 75.76/63.96 & 83.45/56.44 & 81.28/57.56 & \underline{92.29}/\textbf{43.78} & 89.23/43.92 \\
\rowcolor{gray!15} ROSS-GEN & \underline{76.99}/\underline{63.24} & \underline{75.55}/\underline{64.19} & \underline{83.33}/\underline{56.66} & \underline{81.07}/\underline{57.69} & 92.33/\underline{43.73} & \underline{89.19}/\underline{43.94} \\
\rowcolor{gray!15} ROSS-fDBD & \textbf{73.67}/\textbf{64.47} & \textbf{72.10}/\textbf{65.42} & \textbf{80.43}/\textbf{57.13} & \textbf{77.52}/\textbf{58.48} & \textbf{90.68}/42.59 & \textbf{86.30}/\textbf{44.72} \\
\bottomrule
\end{tabular}
}
\end{table*}

In Table~\ref{tab:ross_gen_cifar10_comparison}, we evaluate ROSS ($\sclam$) against three key post-hoc baselines: the median score ($\smed$), the stability score ($\mad$), and their naive ratio ($\smed/\mad$) in order to show the effectiveness of ROSS compared to its individual components.
We report results as FPR95 $\downarrow$ / AUROC $\uparrow$, across near-OOD (CIFAR-100, TIN) and far-OOD (MNIST, SVHN, Texture, Places365) datasets using CIFAR-10 as the in-distribution set and using the base GEN~\cite{liu_gen_2023} OOD score, which PRO~\cite{chen_leveraging_2025} utilised as their best performing baseline.

As expected, $\smed$ performs well overall, especially on far-OOD datasets like MNIST and Places365, due to the effectiveness of the underlying base score on these types of OOD datasets. 

The $\mad$ baseline directly measures score instability. 
It achieves the lowest FPR95 on SVHN, but suffers from poor performance on datasets like Texture, highlighting its inconsistent discrimination power when considered in isolation. 
We hypothesise that some semantically dissimilar, far-OOD samples produce highly stable scores due to network indifference.

The ratio $\smed/\mad$ is a naive combination of both scores by rewarding high score magnitude and low variability. 
It achieves the best FPR95 on MNIST, though its overall performance is poor. 
It is particularly inaccurate on some far-OOD datasets, such as SVHN, where the $\mad$ score is an unreliable discriminator and dominates the overall score.

Our final score, $\sclam$, addresses these limitations by introducing a confidence-aware bonus that activates only when $\smed$ exceeds a learned threshold. 
This design ensures that stability is only rewarded for inputs likely to be ID. 
As shown in Table~\ref{tab:ross_gen_cifar10_comparison}, $\sclam$ consistently ranks among the top scores across all datasets and achieves the best overall average (FPR95: 45.77\%, AUROC: 84.72\%). 
It also maintains the best FPR95 over all datasets and the top AUROC score in five of the six datasets, improving ID vs OOD separation across both near- and far-OOD scenarios.

These results confirm our core idea: by combining a stable score, smoothing to reduce noise, and filtering out low-confidence predictions, ROSS leads to more reliable OOD detection, even under different types of distribution shifts.

\subsection{Evaluation Under Adversarial Attacks}
We evaluate the robustness of ROSS against adversarial perturbations generated by PGD attacks of varying strengths ($\epsilon = 2/255$ to $8/255$), considering both PGD-min and PGD-max configurations. ROSS is applied as a post-processing step to four standard OOD detection methods—MSP, EBO, GEN, and fDBD—and compared to their original counterparts. We also include comparisons with perturbation-based baselines: ODIN and PRO (the latter applied to the best-performing base method, fDBD). Tables~\ref{tab:combined_attacks_cifar10} and~\ref{tab:combined_attacks_cifar100} report averaged FPR95 ($\downarrow$) and AUROC ($\uparrow$) across multiple OOD benchmarks.

\paragraph{ROSS Enhances Baseline Robustness}  
Our theoretical framework hypothesises that ID samples are located within a more stable landscape than OOD points, especially near-OOD points, which reside closer to decision boundaries.
However, rewarding stability on its own can reward OOD points that reside in low-confidence regions.
ROSS mitigates these issues by rewarding stability for high-confidence inputs and thus, ROSS improves the robustness of baseline OOD detection methods.

Our empirical analysis reinforces this hypothesis. On CIFAR-10 under PGD-Max attacks ($\epsilon = 8/255$), MSP’s performance degrades sharply (FPR95/AUROC: 99.99\% / 3.76\%), while ROSS-MSP improves robustness substantially (89.83\% / 52.65\%). Similarly, ROSS-fDBD enhances fDBD’s degraded performance (99.64\% / 17.86\%) to 81.99\% / 58.81\%. These results demonstrate ROSS’s ability to strengthen the adversarial robustness of baseline OOD detectors.

We see similar results on CIFAR-100. 
At $\epsilon=8/255$ (PGD-max), ROSS-GEN improves AUROC from 5.87\% to 43.94\%
, and ROSS-fDBD increases AUROC from 11.02\% to 44.72\%.  
Even simpler methods like MSP see improvements in accuracy, with AUROC increasing from 5.73\% to 37.54\%, further confirming that score stability and thresholding base scores are key to robust OOD detection.

\paragraph{Comparison with Perturbation-based Detectors}  
In this section, we compare ROSS to other perturbation-based OOD detectors, namely ODIN~\cite{dcbe7abf4db64d1b89bf9802585660ed} and PRO~\cite{chen_leveraging_2025}.
As discussed in section~\ref{sec:intro}, both ODIN and PRO leverage perturbations in their scoring functions.
ODIN uses these perturbations to increase ID scores, whereas PRO works to lower OOD scores.

Despite their efficacy being limited to small perturbation magnitudes, their directional optimisation leads to an incidental, but asymmetric, form of robustness. 
Because ODIN pushes scores upward, it provides some robustness against max-attacks that also seek to maximise the score. 
On the other hand, PRO's score-minimising defends against min-attacks. 
However, the directional nature of these methods leaves them fundamentally vulnerable to attacks from the opposing direction. 

For instance, on the CIFAR-100 benchmark (Table~\ref{tab:combined_attacks_cifar100}) at an attack strength of $\epsilon=4/255$, PRO-fDBD records an AUROC of 38.82\% against the PGD-min attack but drops to 28.38\% against the opposing PGD-max attack. 
Similarly, ODIN struggles more against the PGD-max attack, with its AUROC dropping from 31.36\% (PGD-min) to 27.76\% (PGD-max). 
In fact, PRO and ODIN can worsen base scores under certain adversarial conditions. 
This happens because their directional optimisation adversely affects the opposing attacks. 
For example, when facing a PGD-max attack on the CIFAR-100 benchmark (with $\epsilon=4/255$), the baseline fDBD detector achieves a 30.40\% AUROC. However, applying PRO, which is designed to minimise scores, actually degrades performance, resulting in a lower AUROC of 28.38\%. 

ROSS, on the other hand, is not only more robust overall but also avoids this asymmetric vulnerability. 
Under the same attack conditions, ROSS-fDBD records an AUROC of 57.13\%
against PGD-min and 58.48\% 
against PGD-max attacks. 
It is also worth noting that, unlike PRO and ODIN, ROSS is a gradient-free method and quantifies instability directly from random perturbations.

\paragraph{Performance on CIFAR-10 versus CIFAR-100}
We observe that ROSS's performance gains are significantly more pronounced on CIFAR-100 (ID) than on CIFAR-10 (ID). 
We hypothesise this is because the decision boundaries for the 100-class problem are inherently more complex and high-dimensional. 
This complexity makes baseline detectors more susceptible to landscape instability, a vulnerability that methods like PRO-fDBD fail to defend against. 
In contrast, ROSS, by design, directly quantifies and leverages this instability.
\section{Conclusion}
We introduced ROSS, a robust, model-agnostic post-hoc OOD scoring framework that enhances OOD detection under both natural distribution shifts and adversarial perturbations.
Unlike prior work that overlooks robustness, ROSS improves resilience through localised stability, which penalises score volatility under small input perturbations, and confidence-based thresholding, which adapts decisions to model certainty.

While stronger attacks may still benefit from adversarially trained defences such as ALOE or ATOM, ROSS offers a lightweight and easily deployable first-line defence, yielding significantly more stable scores against attacks than baseline methods.
ROSS integrates seamlessly with existing OOD scoring approaches, enhancing robustness without retraining.

Comprehensive experiments demonstrate consistent gains across OOD benchmarks and adversarial settings, achieving up to a 40-point improvement in AUROC over prior perturbation-based methods under PGD-max attacks.
This demonstrates that ROSS provides a practical solution for robust post-hoc OOD detection.

Future work will explore extending ROSS beyond gradient-based attacks to encompass distributional shifts induced by generative models, semantic perturbations, and real-world corruptions, broadening its applicability to more complex and adaptive threat models.
\section*{Acknowledgements} 
Maria Stoica is supported by the UKRI Centre for Doctoral Training in Safe and Trusted Artificial Intelligence (EP/S023356/1).
Abdelrahman Hekal was supported by the European Union’s Horizon research and innovation programme under the NeSy project (grant agreement No. 101070430) while at Imperial College London. 
Alessio Lomuscio is supported by a Royal Academy of Engineering Chair in Emerging Technologies.

{
    \small
    \bibliographystyle{ieeenat_fullname}
    \bibliography{refs_final}
}

\appendix
\clearpage
\setcounter{page}{1}
\maketitlesupplementary
\noindent{The appendix is organised as follows:}
\begin{itemize}
    \item In Section~\ref{sup:large_scale}, we provide results for our proposed post-processing methods on CIFAR-100 and ImageNet.
    \item In Section~\ref{sup:ablation}, we provide ablation studies for hyperparameters $\lambda$, number of samples, and the noise magnitude.
    \item In Section~\ref{sup:alt_scores}, we analyse our post-processor using other OOD baseline scores (MSP, EBO, and fDBD).
    \item In Section~\ref{sup:adaptive_attacks}, we discuss the robustness of ROSS against adaptive and black-box attacks.
    \item In Section~\ref{sup:latency}, we provide an analysis of inference latency and computational overhead.
    \item In Section~\ref{sup:directional}, we expand on the theoretical differences between ROSS and directional perturbation methods.
    \item Finally, in Section~\ref{sup:methodology}, we detail methodological clarifications regarding score direction and validation splits.
\end{itemize}

\section{Benchmark Results on Large-Scale Datasets}\label{sup:large_scale}
In this section, we present additional results for our proposed post-processing methods for out-of-distribution (OOD) detection. In addition to the evaluation on CIFAR-10 presented in the main text, we also present results for CIFAR-100 and ImageNet.

\begin{table*}[p]
 \centering
 \caption{
Analysis of various post-processors using the base \textbf{GEN} score for a model trained on \textbf{CIFAR-100} (ID) and evaluated against various OOD benchmarks. Performance is reported as FPR95 (\%) $\downarrow$ / AUROC (\%) $\uparrow$. Best metric is in \textbf{bold} and second best is \underline{underlined}.
 }
 \label{tab:your_cifar100_results}
 \begin{tabular}{lcccccc|c}
 \toprule
 \textbf{Post-processor} & \multicolumn{2}{c}{\textbf{near-OOD}} & \multicolumn{4}{c|}{\textbf{far-OOD}} & \textbf{Avg.} \\
 \cmidrule(lr){2-3} \cmidrule(lr){4-7}
 & \textbf{CIFAR-10} & \textbf{TIN} & \textbf{MNIST} & \textbf{SVHN} & \textbf{Texture} & \textbf{Places365} & \\
 \midrule
 $\smed$ & \underline{72.65}/\textbf{69.48} & \textbf{67.18}/\textbf{74.08} & \underline{60.88}/\underline{71.92} & \textbf{56.75}/\textbf{75.33} & \textbf{86.36}/\underline{56.88} & \textbf{72.87}/\textbf{71.41} & \textbf{69.45}/\underline{69.85} \\
 $\mad$ & 97.35/41.15 & 98.56/35.58 & 99.29/18.15 & 97.44/35.57 & 99.33/38.18 & 98.16/37.48 & 98.36/34.35 \\
 $\smed/\mad$ & 84.11/66.21 & 80.28/71.27 & \textbf{44.65}/\textbf{83.66} & 72.80/72.40 & 89.11/\textbf{63.66} & 83.24/68.97 & 75.70/\textbf{71.03} \\
 $\sclam$ & \textbf{72.64}/\underline{69.48} & \underline{67.21}/\underline{74.08} & 60.96/71.87 & \underline{56.77}/\underline{75.32} & \underline{86.42}/56.84 & \underline{72.88}/\underline{71.41} & \underline{69.48}/69.83 \\
 \bottomrule
 \end{tabular}
 \end{table*}

\begin{table*}[p]
\centering
\caption{
Analysis of various post-processors using the base \textbf{GEN} score for a ResNet-50 model trained on \textbf{ImageNet} (ID) and evaluated against several OOD benchmarks. 
Performance is reported as FPR95 (\%) $\downarrow$ / AUROC (\%) $\uparrow$. Best results are in \textbf{bold}; second best are \underline{underlined}.
}
\label{tab:imagenet_ross_gen}
\begin{tabular}{lccccc|c}
\toprule
\textbf{Post-processor} & \multicolumn{2}{c}{\textbf{near-OOD}} & \multicolumn{3}{c|}{\textbf{far-OOD}} & \textbf{Avg.} \\
\cmidrule(lr){2-3} \cmidrule(lr){4-6}
& \textbf{SSB-hard} & \textbf{NINCO} & \textbf{iNaturalist} & \textbf{Textures} & \textbf{OpenImage-O} & \\
\midrule
$\smed$ & $\underline{77.40}/\underline{71.63}$ & $\underline{52.46}/\underline{82.60}$ & $\underline{29.89}/\underline{91.24}$ & $\underline{48.51}/\textbf{87.02}$ & $\underline{36.69}/\underline{88.41}$ & $\underline{48.99}/\underline{84.18}$ \\
$\mad$ & $98.01/48.51$ & $99.07/48.98$ & $99.70/32.79$ & $99.94/27.94$ & $99.65/33.96$ & $99.27/38.44$ \\
$\smed/\mad$ & $91.35/60.08$ & $92.46/63.20$ & $82.96/78.37$ & $84.78/79.58$ & $85.08/76.61$ & $87.53/71.57$ \\
$\sclam$ & $\textbf{76.89}/\textbf{71.67}$ & $\textbf{51.65}/\textbf{82.77}$ & $\textbf{29.61}/\textbf{91.29}$ & $\textbf{48.48}/\underline{86.90}$ & $\textbf{36.49}/\textbf{88.43}$ & $\textbf{48.62}/\textbf{84.21}$ \\
\bottomrule
\end{tabular}
\end{table*}

On the CIFAR-100 benchmark, the stability measure ($\mad$) alone provides poor separation between in-distribution (ID) and out-of-distribution (OOD) data. 
At the same time, the median-smoothed GEN score ($\smed$) is far more effective. 
A naive combination, represented by the $\smed$/$\mad$ score, significantly degrades detection performance compared to using the median score by itself. 

For the ImageNet benchmark, we see a similar trend. 
The $\mad$ and $\smed$/$\mad$ scores provide poor ID-OOD separation.
Both $\smed$ and $\sclam$ provide the best results, with $\sclam$ narrowly winning out.

These findings are consistent with insights from PRO~\cite{chen_leveraging_2025}, which found that for large-scale models, the score instability of ID inputs can become indistinguishable from that of OOD inputs. Despite this challenge, our proposed method, ROSS, successfully integrates the stability metric without this penalty, resulting in almost no degradation compared to the strong baseline median scores. 

\section{Ablation Studies}\label{sup:ablation}

In this section, we present a series of ablation studies to examine the impact of key components and hyperparameters in our proposed ROSS framework.

\subsection{Sensitivity to Hyperparameter $\lambda$}

We present experiments with different values for the stability weighting parameter $\lambda$. The analysis is performed on models trained on CIFAR-10, CIFAR-100 and ImageNet.

\begin{table*}[ht!]
\centering
\caption{
Analysis of the ROSS-GEN post-processor on \textbf{CIFAR-10} with different lambda ($\lambda$) values. Performance is reported as FPR95 (\%) $\downarrow$ / AUROC (\%) $\uparrow$. Best metric is in \textbf{bold} and second best is \underline{underlined}.
}
\label{tab:ross_lambda_variants}
\begin{tabular}{lcccccc|c}
\toprule
\textbf{$\lambda$} & \textbf{CIFAR-100} & \textbf{TIN} & \textbf{MNIST} & \textbf{SVHN} & \textbf{Texture} & \textbf{Places365} & \textbf{Avg.} \\
\midrule
0.005 & 58.75/83.20 & 49.90/85.80 & 49.83/84.68 & 57.42/76.89 & 52.04/84.72 & 42.14/88.53 & 51.68/83.97 \\
0.01 & 55.74/83.46 & 48.51/86.02 & 48.33/84.87 & 54.86/77.42 & 51.23/\textbf{84.80} & 41.39/88.69 & 50.01/84.21 \\
0.02 & 53.35/83.71 & 46.66/86.21 & 46.69/85.03 & 51.57/77.96 & \textbf{49.98}/\underline{84.75} & 40.32/88.82 & 48.09/84.41 \\
0.05 & \underline{49.90}/\underline{84.09} & \underline{44.26}/\underline{86.53} & \underline{44.28}/\textbf{85.23} & \underline{46.89}/\underline{78.80} & \underline{50.33}/84.59 & \underline{38.98}/\underline{89.05} & \underline{45.77}/\underline{84.72} \\
0.1 & \textbf{48.43}/\textbf{84.26} & \textbf{43.53}/\textbf{86.65} & \textbf{43.70}/\underline{85.18} & \textbf{44.88}/\textbf{79.21} & 51.13/84.26 & \textbf{38.44}/\textbf{89.15} & \textbf{45.02}/\textbf{84.79} \\
\bottomrule
\end{tabular}
\end{table*}

\begin{table*}[ht!]
\centering
\caption{
Analysis of the ROSS-GEN post-processor on \textbf{CIFAR-100} with different lambda ($\lambda$) values. Performance is reported as FPR95 (\%) $\downarrow$ / AUROC (\%) $\uparrow$. Best metric is in \textbf{bold} and second best is \underline{underlined}.
}
\label{tab:ross_lambda_variants_cifar100}
\begin{tabular}{lcccccc|c}
\toprule
\textbf{$\lambda$} & \textbf{CIFAR-10} & \textbf{TIN} & \textbf{MNIST} & \textbf{SVHN} & \textbf{Texture} & \textbf{Places365} & \textbf{Avg.} \\
\midrule
0.005 & 72.68/69.55 & \textbf{66.79}/\textbf{74.15} & \textbf{60.74}/\textbf{71.95} & \textbf{56.70}/75.37 & 86.66/56.91 & 72.81/\textbf{71.46} & \textbf{69.40}/\textbf{69.90} \\
0.01 & 72.64/69.48 & 67.21/74.08 & 60.96/\underline{71.87} & 56.77/75.32 & 86.42/56.84 & 72.88/71.41 & 69.48/69.83 \\
0.02 & 72.68/69.55 & \textbf{66.79}/74.14 & \underline{60.85}/\underline{71.87} & \textbf{56.70}/75.37 & 86.80/56.84 & 72.83/\textbf{71.46} & \underline{69.44}/\underline{69.87} \\
0.05 & \textbf{72.33}/\underline{69.99} & 67.23/74.12 & 69.00/64.53 & 57.94/\textbf{76.91} & \textbf{85.13}/\textbf{58.42} & \textbf{72.36}/71.19 & 70.67/69.19 \\
0.1 & \underline{72.35}/\textbf{70.10} & 67.28/\textbf{74.16} & 68.98/63.80 & 58.02/\underline{76.87} & \underline{85.21}/\underline{58.10} & \underline{72.45}/71.24 & 70.72/69.05 \\
\bottomrule
\end{tabular}
\end{table*}

\begin{table*}[ht!]
\centering
\caption{
Analysis of the ROSS-GEN post-processor on \textbf{ImageNet} with different lambda ($\lambda$) values. Performance is reported as FPR95 (\%) $\downarrow$ / AUROC (\%) $\uparrow$. Best metric is in \textbf{bold} and second best is \underline{underlined}.
}
\label{tab:ross_lambda_variants_imagenet}
\begin{tabular}{lccccc|c}
\toprule
\textbf{$\lambda$} & \textbf{SSB-Hard} & \textbf{NINCO} & \textbf{iNaturalist} & \textbf{Textures} & \textbf{OpenImage-O} & \textbf{Avg.} \\
\midrule
0.005 & 77.24/71.63 & 52.35/82.60 & 29.77/91.24 & \textbf{48.34}/\textbf{87.03} & 37.05/88.41 & 48.95/84.18 \\
0.01 & 77.18/71.64 & 52.25/82.62 & 29.52/91.25 & \underline{48.42}/\underline{86.99} & 36.78/88.42 & 48.83/84.18 \\
0.02 & 77.07/71.65 & 51.88/82.65 & \underline{29.42}/91.25 & \underline{48.43}/\underline{86.98} & 36.81/\textbf{88.43} & 48.72/84.19 \\
0.05 & \underline{76.89}/\underline{71.67} & \underline{51.65}/\underline{82.77} & 29.61/\underline{91.29} & 48.48/86.90 & \underline{36.49}/\textbf{88.43} & \underline{48.62}/\textbf{84.21} \\
0.1 & \textbf{76.75}/\textbf{71.68} & \textbf{50.12}/\textbf{82.89} & \textbf{29.11}/\textbf{91.31} & 48.70/86.78 & \textbf{36.34}/\textbf{88.44} & \textbf{48.20}/\textbf{84.22} \\
\bottomrule
\end{tabular}
\end{table*}

The selection of the hyperparameter $\lambda$ in the ROSS score is a trade-off between the effect of the median score and stability. For the CIFAR-10 model, a $\lambda$ value of $0.01$ or $0.02$ achieves the best overall average performance. 
For CIFAR-100, a smaller value of $\lambda = 0.005$ provides the best average results, primarily due to the median score providing a stronger separation in these models. However, varying values of $\lambda$ have little effect on the final results. 
For ImageNet, we see that a larger $\lambda$ of 0.1 produces the best results.
As with CIFAR-100, the choice of $\lambda$ does not change the results drastically.

\subsection{Number of Samples}

We evaluate the robustness of the ROSS-fDBD score against PGD-min and PGD-max adversarial attacks on a CIFAR-10 model. We analyse the impact of the number of samples ($N$) used to compute the score and the magnitude of internal noise ($\sigma$).

\begin{table*}[ht!]
\centering
\caption{OOD detection robustness vs. number of samples ($N$) for noise $\sigma=0.1$ and $\lambda = 0.05$. Results are averaged over all benchmarks for a \textbf{CIFAR-10} model under PGD attack and reported as FPR95 (\%) $\downarrow$ / AUROC (\%) $\uparrow$.}
\label{tab:table_n_ablation_study}
\resizebox{\textwidth}{!}{%
\begin{tabular}{l|cc|cc|cc}
\toprule
\textbf{Samples ($N$)} & \multicolumn{2}{c|}{\textbf{$\epsilon=2/255$}} & \multicolumn{2}{c|}{\textbf{$\epsilon=4/255$}} & \multicolumn{2}{c}{\textbf{$\epsilon=8/255$}} \\
\cmidrule(lr){2-3} \cmidrule(lr){4-5} \cmidrule(lr){6-7} & \textbf{PGD-Min} & \textbf{PGD-Max} & \textbf{PGD-Min} & \textbf{PGD-Max} & \textbf{PGD-Min} & \textbf{PGD-Max} \\
\midrule
5 & 59.48/78.34 & 60.09/78.79 & 71.52/71.63 & 69.94/72.39 & 88.27/56.98 & 87.98/54.83\\
10 & 58.33/78.58 & 58.64/79.03 & 70.68/71.86 & 68.45/72.64 & 88.03/57.09 & 87.76/54.79\\
25 & 58.42/78.75 & 58.80/78.78 & 70.83/71.95 & 70.35/71.62 & 88.30/56.85 & 88.08/54.32\\
50 & 58.07/78.76 & 57.58/79.27 & 70.53/71.95 & 67.42/72.83 & 88.17/56.85 & 87.66/54.79\\
100 & 58.05/78.79 & 57.59/79.32 & 70.61/71.95 & 66.62/73.27 & 88.24/56.78 & 87.75/54.75\\
\bottomrule
\end{tabular}%
}
\end{table*}

\begin{table*}[ht!]
\centering
\caption{OOD detection robustness vs. noise magnitude ($\sigma$) for sample size $N=25$. Results are averaged over all benchmarks for a \textbf{CIFAR-10} model under PGD attack using GEN and reported as FPR95 (\%) $\downarrow$ / AUROC (\%) $\uparrow$.}
\label{tab:robustness_vs_noise}
\resizebox{\textwidth}{!}{%
\begin{tabular}{l|c|cc|cc|cc}
\toprule
\textbf{Noise Magnitude} & \textbf{No Attack} & \multicolumn{2}{c|}{\textbf{$\epsilon=2/255$}} & \multicolumn{2}{c|}{\textbf{$\epsilon=4/255$}} & \multicolumn{2}{c}{\textbf{$\epsilon=8/255$}} \\
\cmidrule(lr){3-4} \cmidrule(lr){5-6} \cmidrule(lr){7-8}
& ($\epsilon=0$) & Min & Max & Min & Max & Min & Max\\
\midrule
 0.025 & 32.19/91.15 & 58.27/79.34 & 65.61/77.65 & 77.81/65.98 & 84.74/55.65 & 94.35/44.82 & 95.68/18.94 \\
 0.05 & 36.07/89.55 & 56.45/80.69 & 53.17/79.34 & 74.02/71.11 & 71.12/66.65 & 92.25/54.90 & 85.41/43.85 \\
 0.1 & 45.77/84.72 & 58.42/78.75 & 58.80/78.78 & 70.83/71.95 & 70.35/71.62 & 88.30/56.85 & 88.08/54.32\\
 0.25 & 82.11/58.84 & 82.95/58.87 & 82.96/57.71 & 83.90/57.41 & 83.69/57.21 & 85.60/55.80 & 83.90/50.46 \\
\bottomrule
\end{tabular}%
}
\end{table*}

While smaller sample sizes ($N=5, 10$) occasionally yield better scores, these results are attributed to the high variance and noise inherent in limited sampling. 
As $N$ increases, the metrics converge, reflecting a more accurate and stable estimation of the model's true robustness against adversarial perturbations. 
Consequently, we adopt $N=25$ as it represents an optimal trade-off, providing the stability observed at higher sample counts ($N=100$) while maintaining manageable runtime costs.

\subsection{Noise Magnitude: The Robustness-Accuracy Trade-off}
A key advantage of ROSS is that it allows for an explicit trade-off between performance on clean data and adversarial robustness, controlled by the noise magnitude hyperparameter $\sigma$. 
As shown in the data presented in Table~\ref{tab:robustness_vs_noise} and visualised for PGD-Max attacks in Figure~\ref{fig:noise-tradeoff}, a small noise value ($\sigma=0.025$) achieves the highest AUROC on unattacked inputs (91.15\%) but collapses under strong attacks, dropping to just 18.94\% AUROC against a PGD-max attack with $\epsilon=8/255$. A larger noise magnitude ($\sigma=0.25$) boosts adversarial robustness, maintaining an AUROC of 50.46\% under the same strong attack, but it degrades performance on clean data (58.84\% AUROC). 
In our main body of experiments, we use $\sigma=0.1$, which offers a practical balance, sacrificing a minimal amount of performance on clean inputs to achieve a substantial improvement in adversarial robustness over the lower-noise setting. 

\section{Performance with Alternative Base Scores}\label{sup:alt_scores}
We extend the analysis of ROSS beyond the GEN base score. This section presents results for ROSS and the baseline score when applied to three other widely-used OOD detection scores: Maximum Softmax Probability (MSP), Energy-based OOD detection (EBO), and Fast Decision Boundary OOD detector (fDBD). 
For each base score, we again evaluate the performance of the median score ($S_{\text{med}}$), the stability score ($\sigma_{\text{med}}$), their naive ratio, and our final proposed score ($S_{\text{ROSS}}$) on a model trained on CIFAR-10 (Tables \ref{tab:our_scores_msp}-\ref{tab:our_scores_fdbd}).

While median smoothing ($\smed$) performs competitively, marginally outperforming ROSS on the fDBD benchmark, they exhibit sensitivity to the underlying score distribution. 
This instability is most evident on the GEN score (Table \ref{tab:ross_gen_cifar10_comparison}), where ROSS achieves an average FPR95 of 45.77\%, significantly outperforming $\smed$ (53.79\%) by a margin of 8\%. 
Similarly, on EBO, ROSS improves upon $\smed$ by over 7\% (Table \ref{tab:our_scores_ebo}). 
ROSS mitigates this sensitivity by adapting to the noise profile of each score. 
Although $\smed$ achieves a slight edge in isolated cases, ROSS delivers consistently optimal or near-optimal metrics across all tested base scores (MSP, EBO, fDBD, and GEN). 
This cross-score stability establishes ROSS as the more robust, agnostic choice for OOD detection.

\begin{table*}[ht!]
\centering
\caption{
Analysis of proposed post-processors using base \textbf{MSP} score. The model is trained on \textbf{CIFAR-10} (ID) and evaluated against various OOD benchmarks. Performance is reported as FPR95 (\%) $\downarrow$ / AUROC (\%) $\uparrow$. Best metric is in \textbf{bold} and second best is \underline{underlined}
}
\label{tab:our_scores_msp}
\begin{tabular}{lcccccc|c}
\toprule
\textbf{Post-processor} & \multicolumn{2}{c}{\textbf{near-OOD}} & \multicolumn{4}{c|}{\textbf{far-OOD}} & \textbf{Avg.} \\
\cmidrule(lr){2-3} \cmidrule(lr){4-7}
& \textbf{CIFAR-100} & \textbf{TIN} & \textbf{MNIST} & \textbf{SVHN} & \textbf{Texture} & \textbf{Places365} & \\
\midrule
$\smed$ & 53.80/\underline{83.09} & 46.56/\underline{85.30} & 48.38/\textbf{84.24} & 48.58/80.74 & \textbf{45.75}/\textbf{84.90} & 40.46/\textbf{87.23} & 47.25/\textbf{84.25} \\
$\mad$ & \textbf{46.94}/82.10 & \underline{43.17}/83.93 & \underline{45.62}/80.29 & \textbf{40.14}/\textbf{82.16} & \underline{48.00}/80.04 & \underline{40.00}/85.14 & \textbf{43.98}/82.28 \\
$\smed/\mad$ & \underline{46.95}/82.90 & \textbf{43.16}/84.90 & 45.63/81.74 & \underline{40.16}/\underline{82.04} & 48.00/81.50 & 40.00/86.46 & \underline{43.98}/83.26 \\
$\sclam$ & 46.95/\textbf{83.48} & 43.17/\textbf{85.48} & \textbf{45.60}/\underline{83.46} & 40.19/81.97 & 48.00/\underline{83.56} & \textbf{39.99}/\underline{87.17} & 43.98/\underline{84.19} \\
\bottomrule
\end{tabular}
\end{table*}

\begin{table*}[ht!]
\centering
\caption{
Analysis of proposed post-processors using the base \textbf{EBO} score. The model is trained on \textbf{CIFAR-10} (ID) and evaluated against various OOD benchmarks. Performance is reported as FPR95 (\%) $\downarrow$ / AUROC (\%) $\uparrow$. Best metric is in \textbf{bold} and second best is \underline{underlined}
}
\label{tab:our_scores_ebo}
\begin{tabular}{lcccccc|c}
\toprule
\textbf{Post-processor} & \multicolumn{2}{c}{\textbf{near-OOD}} & \multicolumn{4}{c|}{\textbf{far-OOD}} & \textbf{Avg.} \\
\cmidrule(lr){2-3} \cmidrule(lr){4-7}
& \textbf{CIFAR-100} & \textbf{TIN} & \textbf{MNIST} & \textbf{SVHN} & \textbf{Texture} & \textbf{Places365} & \\
\midrule
$\smed$ & 64.98/\underline{85.72} & 54.54/\underline{88.30} & \underline{31.63}/\underline{92.29} & 63.97/80.97 & \underline{52.81}/\underline{88.30} & 50.65/\underline{89.38} & 53.10/\underline{87.49} \\
$\mad$& 65.00/\underline{82.47} & 55.88/\underline{85.28} & \underline{52.44}/\underline{84.27} & 64.57/73.86 & \underline{57.61}/\textbf{84.16} & \underline{45.96}/\underline{88.32} & 56.91/\underline{83.06} \\
$\smed/\mad$ & \underline{52.83}/76.59 & \underline{48.90}/78.14 & 54.55/69.28 & \underline{45.59}/\textbf{78.89} & 80.30/65.39 & 46.98/79.20 & \underline{54.86}/74.58 \\
$\sclam$  & \textbf{48.63}/\textbf{84.17} & \textbf{43.23}/\textbf{86.66} & \textbf{44.14}/\textbf{84.83} & \textbf{45.53}/\underline{78.34} & \textbf{55.81}/\underline{83.21} & \textbf{38.34}/\textbf{89.33} & \textbf{45.95}/\textbf{84.42} \\
\bottomrule
\end{tabular}
\end{table*}

\begin{table*}[ht!]
 \centering
 \caption{
Analysis of proposed post-processors using the base \textbf{fDBD} score. The model is trained on \textbf{CIFAR-10} (ID) and evaluated against various OOD benchmarks. Performance is reported as FPR95 (\%) $\downarrow$ / AUROC (\%) $\uparrow$. Best metric is in \textbf{bold} and second best is \underline{underlined}
 }
 \label{tab:our_scores_fdbd}
 \begin{tabular}{lcccccc|c}
 \toprule
 \textbf{Post-processor} & \multicolumn{2}{c}{\textbf{near-OOD}} & \multicolumn{4}{c|}{\textbf{far-OOD}} & \textbf{Avg.} \\
 \cmidrule(lr){2-3} \cmidrule(lr){4-7}
 & \textbf{CIFAR-100} & \textbf{TIN} & \textbf{MNIST} & \textbf{SVHN} & \textbf{Texture} & \textbf{Places365} & \\
 \midrule
 $\smed$ & 49.19/\underline{84.72} & 42.44/\underline{87.32} & \textbf{36.40}/\textbf{87.25} & \textbf{36.49}/\textbf{84.05} & \textbf{41.84}/\textbf{87.20} & \textbf{34.31}/\textbf{90.22} & \textbf{40.11}/\textbf{86.79} \\
 $\mad$ & 47.61/77.73 & 44.19/79.00 & 48.41/70.57 & 42.09/77.63 & 63.39/69.09 & 41.83/79.60 & 47.92/75.60 \\
 $\smed/\mad$ & \underline{46.03}/81.79 & \underline{42.00}/83.99 & 45.23/76.98 & 39.20/79.95 & 56.45/75.76 & 38.56/86.02 & 44.58/80.75 \\
 $\sclam$ & \textbf{45.91}/\textbf{85.10} & \textbf{40.57}/\textbf{87.58} & \underline{38.65}/\underline{86.72} & \underline{36.51}/\underline{83.87} & \underline{45.41}/\underline{86.41} & \underline{34.79}/\underline{90.19} & \underline{40.31}/\underline{86.64} \\
 \bottomrule
 \end{tabular}
 \end{table*}

 \section{Robustness Against Adaptive Attacks}\label{sup:adaptive_attacks}
To thoroughly evaluate the robustness of ROSS and verify that our results are not over-optimistic, we consider adaptive attacks such as Expectation over Transformation (EOT)~\cite{Athalye18}. In our standard evaluation, we directly target the base scoring function. This provides the most precise gradient signal, as the goal is to shift the score distribution sufficiently to alter the robust statistic~\cite{Cohen19}.

As demonstrated by Gao et al.~\cite{Gao22}, EOT offers no advantage over standard Projected Gradient Descent (PGD) attacks unless the noise level is substantial ($\sigma \ge 0.5$).
For ROSS, which operates effectively at a relatively low noise magnitude ($\sigma=0.1$), standard attacks are often optimal.
This is because base gradients provide precise descent directions, whereas EOT relies on high-variance estimates \cite{Salman19}.
Furthermore, it is critical to note that smoothing within ROSS does not result in gradient masking; rather, it is a derived statistical property of the landscape. 

Empirically, when we ran adaptive EOT attacks (using CIFAR-10 as ID, $N=5$, $\epsilon=4/255$) explicitly targeting ROSS, they proved weaker than standard PGD applied to the base post-processor (resulting in approximately +3\% AUROC and -7\% FPR95 compared to PGD). We also investigated black-box attack methods and found that, within the standard adversarial threat model ($\epsilon \le 8/255$), black-box attacks were unable to reliably degrade the ROSS score.
Successful degradation required perturbation magnitudes significantly higher than the standard radius. 

\section{Inference Latency Analysis}\label{sup:latency}
While ROSS requires $N$ forward passes ($N=25$) to compute the stability statistic, it is important to note that these evaluations are entirely independent and highly parallelisable. On modern GPU architectures, the $N$ perturbed inputs are naturally processed in a single batch, avoiding sequential bottlenecks. 

We measured the wall-clock inference time on an NVIDIA RTX 4060 Ti using 100 CIFAR-10 images. The base MSP score requires approximately $9.6$ ms/img, while ROSS ($N=25$) takes approximately $11.43$ ms/img. Due to parallelisation, this represents a minor latency increase of only $\sim1.2\times$, rather than a naive $25\times$ scaling. Furthermore, as shown in our ablation studies (Table \ref{tab:table_n_ablation_study}), ROSS remains highly robust even at $N=5$ (where AUROC drops by $<0.5\%$ compared to $N=25$). Therefore, for users with capable GPUs, the latency difference introduced by ROSS is negligible for practical deployments.

\section{Directional Perturbation vs. Non-Directional Stability}\label{sup:directional}
A fundamental distinction between ROSS and prior perturbation-based methods (such as ODIN and PRO) lies in directionality versus stability. Existing methods rely heavily on directional gradient optimisation—specifically ascending or descending scores to push representations apart. This creates an asymmetric vulnerability: optimising an input for one specific direction (e.g., minimising the score) frequently leaves the model acutely vulnerable to the opposite attack direction (e.g., maximising the score). For instance, as shown in Table 3 of the main text, PRO-fDBD degrades severely under PGD-max attacks.

In contrast, ROSS is fundamentally non-directional. We do not computationally optimise the input toward a specific score threshold. Instead, we measure the statistical stability of the local score landscape using the median and median absolute deviation ($\mad$). This approach provides symmetric robustness, remaining highly effective against both PGD-min and PGD-max attacks, and entirely avoids the reliance on fragile gradient tracking during inference. 

\section{Further Evaluation Details}\label{sup:methodology}

\textbf{Score Direction:} Throughout our evaluations, we strictly adhere to the standard OOD detection convention: High Score = ID (In-Distribution) and Low Score = OOD. Consequently, PGD-min attacks attempt to aggressively lower the score of ID samples (to misclassify them as OOD), whereas PGD-max attempts to artificially raise the score of OOD samples (to misclassify them as ID). As detailed in Algorithm 1 and Equation 2 of the main text, the stability bonus is deliberately added (increasing the final score) only if the sample is already high-confidence (ID-like).

\textbf{Validation Split and Thresholding:} The $S_{95}$ threshold is calculated using the standard validation split containing 10\% held-out ID data, ensuring consistency with established frameworks like OpenOOD. It strictly follows the standard False Positive Rate at 95\% True Positive Rate (FPR95) threshold. 

\textbf{Recalibration:} Recalibration of the $S_{95}$ threshold is only necessary if the definition of the ID task changes significantly (e.g., severe domain shift). Under standard conditions, $S_{95}$ is computed efficiently just once post-training.

\end{document}